\documentclass[11pt]{article}

\usepackage[final]{acl}

\usepackage{times}
\usepackage{latexsym}
\usepackage{makecell}
\usepackage{CJKutf8}
\usepackage{stfloats}
\usepackage{amsmath}
\usepackage{amssymb}
\usepackage{amsfonts}
\usepackage{multirow}
\usepackage{adjustbox}
\usepackage{pifont}
\usepackage{xcolor}
\usepackage{hyperref}
\usepackage{arydshln}
\usepackage{subcaption}
\usepackage{rotating}

\definecolor{lightblue}{RGB}{217, 234, 245}
\definecolor{deepblue}{RGB}{166, 202, 240}
\definecolor{deepdeepblue}{RGB}{45,110,187}
\definecolor{lightyellow}{RGB}{255, 255, 153}
\definecolor{deepgreen}{RGB}{0, 88, 42}

\usepackage[T1]{fontenc}

\usepackage[utf8]{inputenc}

\usepackage{microtype}

\usepackage{inconsolata}

\title{BiMax: Bidirectional MaxSim Score for Document-Level Alignment}

\author{Xiaotian Wang$^{1,2}$ \ \ \ \     Takehito Utsuro$^1$ \ \ \ \ Masaaki Nagata$^3$ \\
       $^1$University of Tsukuba \quad $^2$University of Tokyo \\
       $^3$NTT Communication Science Laboratories, NTT Corporation, Japan \\
       $^{1,2}$\texttt{wangxiaotian1999@\_outlook.com},  
       $^1$\texttt{utsuro@\_iit.tsukuba.ac.jp} \\
       $^3$\texttt{masaaki.nagata@\_ntt.com}
       }

\begin{document}
\maketitle
\begin{abstract}

Document alignment is necessary for the hierarchical mining~\citep{banon-etal-2020-paracrawl, morishita-etal-2022-jparacrawl}, 
which aligns documents across source and target languages within the same web domain.
Several high-precision sentence embedding-based methods have been developed,
such as TK-PERT \citep{thompson-koehn-2020-exploiting} 
and Optimal Transport~(OT)~\citep{clark-etal-2019-sentence, el-kishky-guzman-2020-massively}. 
However, given the massive scale of web mining data,
both accuracy and speed must be considered. 
In this paper, we propose a cross-lingual \textbf{Bi}directional \textbf{Max}sim score~(BiMax) 
for computing doc-to-doc similarity, 
to improve efficiency compared to the OT method.
Consequently, on the WMT16 bilingual document alignment task,
BiMax attains accuracy comparable to OT with an approximate 100-fold speed increase.
Meanwhile, we also conduct a comprehensive analysis to
investigate the performance of current state-of-the-art multilingual sentence embedding models.
All the alignment methods in this paper are publicly available
as a tool called \textit{EmbDA}\footnote{
\url{https://github.com/EternalEdenn/EmbDA}
}.

\end{abstract}

\section{Introduction}
\label{sec:intro}
Document alignment is the task of finding parallel document pairs, 
which are identified as translations of each other, within a collection of documents.
It is mainly employed as a preparatory stage within hierarchical mining for parallel sentence pair curation~\citep{banon-etal-2020-paracrawl, morishita-etal-2022-jparacrawl, nagata-etal-2024-japarapat}, 
seeking to enhance pair quality~\citep{sloto-etal-2023-findings, steingrimsson-2023-sentence} by restricting sentence alignment in high-precision aligned document pairs.
With recent advances in document-level machine translation~\citep{sun-etal-2022-rethinking, wang-etal-2023-document-level, wang-etal-2024-benchmarking, pal-etal-2024-document}, 
document alignment has also become a viable strategy for developing high-quality parallel document pairs~\citep{suryanarayanan2024pralekha}.

There are four mainstream approaches:
URL matching~\cite{germann-2016-bilingual, papavassiliou-etal-2016-ilsp},
bilingual lexicon~\cite{azpeitia-etchegoyhen-2016-docal, medved-etal-2016-english},
machine translation~\cite{dara-lin-2016-yoda, buck-koehn-2016-quick},
sentence embedding~\cite{clark-etal-2019-sentence, thompson-koehn-2020-exploiting, el-kishky-guzman-2020-massively}.

\citet{wang-etal-2024-document} proposed the Overlapping Fixed-Length Segmentation~(OFLS)
as an alternative to Sentence-based Segmentation~(SBS) for generating embeddings.
When applied to Mean-Pool, TK-PERT~\cite{thompson-koehn-2020-exploiting},
and OT~\cite{clark-etal-2019-sentence, el-kishky-guzman-2020-massively}, 
this strategy led to both speed and accuracy improvements. 
Among these methods,
OT achieves the highest recall in the WMT16 bilingual document alignment 
shared task~\citep{buck-koehn-2016-findings} based on LaBSE~\citep{feng-etal-2022-language}. 
However, the computation of OT inherently involves an optimization process, 
necessitating multiple iterative operations.
This results in high computational complexity, limiting its performance in speed. 

Thus, we propose the \textbf{Bi}directional \textbf{Max}Sim score~(BiMax),
which matches the maximum similarity between a given segment and the opposed segment collection
and then sums and averages the similarity scores. 
The implementation is computationally efficient,
requiring only a single similarity matrix computation followed by two max-pooling operations.
This idea is inspired by the MaxSim Score in ColBERT~\cite{omar-colbert, santhanam-etal-2022-colbertv2}, 
which uses a late interaction mechanism to reduce the query-passage
computational cost
by calculating only the maximum similarity for each query token relative to the tokens in the passage. 
We extend this score to the sentence level and make it bidirectional.

Additionally, we evaluate combinations of state-of-the-art embedding models~(i.e., models that perform well in tasks such as bitext mining and STS) 
with various segmentation strategies and document alignment methods 
on the small-scale Ja-En MnRN dataset~\cite{wang-etal-2024-document}, 
aiming to find suitable models and methods for different scenarios.
Meanwhile, 
we make a modest attempt to examine the performance of different methods on low-resource languages.
Finally,
we build a downstream MT benchmark\footnote{
This benchmark is offered solely as a reference rather than a definitive proposal, 
thus we include it only in Appendix~\ref{sec:mt_bench}.
}
to assess the impact of document alignment based on the WMT23 Parallel Data Curation task, 
and the construction process is comprehensively and transparently recorded in Appendix~\ref{sec:mt_bench}.

\vspace{-0.15cm}
\section{Related Work}
\label{sec:related}
Currently, there are four mainstream approaches to document alignment.
The first involves simply calculating similarity based on the URLs of the 
documents~\cite{germann-2016-bilingual, papavassiliou-etal-2016-ilsp}. 
The second uses a bag-of-words or bag-of-ngrams representation of the document contents, 
leveraging a bilingual lexicon for computation~\cite{azpeitia-etchegoyhen-2016-docal, medved-etal-2016-english}. 
The third entails translating documents into the same language, 
followed by similarity calculations using ngram-based metrics~(e.g., BLEU)~\cite{dara-lin-2016-yoda, buck-koehn-2016-quick}. 
The fourth utilizes multilingual pre-trained embedding models to map documents into a shared vector space, where similarity is determined by the distances between 
vectors~\cite{clark-etal-2019-sentence, thompson-koehn-2020-exploiting, el-kishky-guzman-2020-massively}.
In the WMT16 bilingual document alignment shared task~\citep{buck-koehn-2016-findings}, 
numerous techniques and systems were proposed. 
However, due to the limitations of technology at the time, 
all efforts focused on the first three approaches, 
with no exploration of embedding-based methods.

With the development of pre-trained multilingual sentence embedding models~\citep{artetxe-schwenk-2019-massively,feng-etal-2022-language},
which map sentences from different languages into a shared multilingual vector space,
cross-lingual bitext mining has become feasible.
This progress also facilitates representing documents using segment embeddings 
and computing doc-pair similarities via vector-based methods.

\citet{thompson-koehn-2020-exploiting} introduced TK-PERT,
a method that assigns weights to sentences using regionally emphasized windows 
derived from a modified PERT distribution~\cite{vose-2000-pert}
to form document feature vectors. 
Optimal Transport~(OT) was also applied in cross-lingual document alignment, 
evolving from the word level with Word Movers' Distance~(WMD)~\cite{pmlr-v37-kusnerb15} 
to the sentence level with Sentence Movers' Distance~SMD) and
Greedy Movers' Distance~(GMD)~\cite{clark-etal-2019-sentence, el-kishky-guzman-2020-massively}.
Building on GMD, 
\citet{ref1} et al. employed a new weighting strategy using bilingual lexicons, 
further enhancing alignment accuracy in low-resource languages.
\citet{wang-etal-2024-document} proposed OFLS 
instead of SBS for the embedding step.
However, their work is limited to using only the LaBSE model 
and does not explore new document alignment methods.

\vspace{-0.15cm}
\section{Method}
\label{sec:method}
Unlike MaxSim utilized in ColBERT~\cite{omar-colbert, santhanam-etal-2022-colbertv2},
which uses the query's hidden word embeddings to search for the most similar token in the passage undirectionally,
we apply it to sentence-level as the Bidirectional MaxSim Score~(BiMax),
introducing the following key modifications:
\( (1) \) transforming from monolingual to cross-lingual,
\( (2) \) shifting from word-level embeddings to sentence-level embeddings,
and
\( (3) \) moving from one-sided maximum similarity matching to 
a bidirectional approach.

\vspace{-0.15cm}
\subsection{Bidirectional MaxSim Score}
We define the source / target document set as  $\mathcal{D}_S$ and $\mathcal{D}_T$,
and adopt a 2-stage approach to consider the 
$\mathcal{D}_S \times \mathcal{D}_T$ possible document pairs:
\vspace{-0.2cm}

\begin{enumerate}
    \item \textbf{Candidate Generation}: We first use Mean-Pool or TK-PERT method to 
        generate a single feature vector for each document, 
        and then employ Faiss Search~\citep{johnson2019billion} 
        to retrieve $K$ target documents as potential matches for each source document.
    \vspace{-0.2cm}
    \item \textbf{Candidate Re-ranking}: We re-rank the $\mathcal{D}_S \times K$ pairs 
        using a more accurate but slower and sometimes more memory-intensive scoring method, 
        such as OT and our proposed BiMax. 
\end{enumerate}

\vspace{-0.2cm}
Let $s_i$ for $ i \in \{0, ..., N_S - 1\}$ be the $N_S$ segments in a given source document $S$ 
and $t_j$ for $ j \in \{0, ..., N_T - 1\}$ be the $N_T$ segments in a given target document $T$.
The BiMax Score is defined as:
\vspace{-1.0cm}

\begin{subequations}
\small
\begin{align}
&\mathrm{MaxSim}(S, T) = \frac{1}{N_S}\sum_{i=1}^{N_S} \mathop{\mathrm{max}}\limits_{t \in T} \mathrm{Sim}(s_i, t) \\
&\mathrm{BiMax}(S, T) = \frac{1}{2} (\mathrm{MaxSim}(S, T)+\mathrm{MaxSim}(T, S)) \label{eq:bimax}
\end{align}
\end{subequations}
\vspace{-0.4cm}

where $\mathrm{Sim}(s,j)$ represents the similarity score.
In this work, we use a pre-trained multilingual sentence embedding model to map the source segment $s$
and the target segment $t$
into the same vector space, producing embeddings $E_s$ and $E_t$,
and then adopting their cosine similarity $cos(E_s, E_t)$.

\begin{table*}[ht]
\centering
\begin{adjustbox}{width=0.86\textwidth}
\begin{tabular}{|ll|cccc|}
\hline
\multicolumn{2}{|l|}{Strategies \& Models}
& \href{https://huggingface.co/setu4993/LaBSE}{LaBSE}
& \href{https://huggingface.co/sentence-transformers/distiluse-base-multilingual-cased-v2}{\makecell[c]{distiluse-base-\\ multi-cased-v2 }} 
& \href{https://huggingface.co/BAAI/bge-m3}{BGE M3}
& \href{https://huggingface.co/jinaai/jina-embeddings-v3}{jina-embed-v3}   \\ 
\hline
\multicolumn{6}{|l|}{\textbf{Common Info.~(Source / Target)}} \\
\hline
\multicolumn{2}{|l}{Document Num.} & \multicolumn{4}{c|}{232 / 931} \\
\multicolumn{2}{|l}{Total Sentence Num.} & \multicolumn{4}{c|}{4,746 / 57,032} \\
\multicolumn{2}{|l}{Gold Pairs} & \multicolumn{4}{c|}{263\footnotemark } \\
\hline
\multicolumn{2}{|l}{Total Document Token Num.} & 0.50M / 3.34M & 0.53M / 3.68M & 0.43M / 3.68M & 0.43M / 3.68M \\
\multicolumn{2}{|l}{Average Sentence Token Num.} & 105.17 / 58.55 & 111.27 / 64.49 & 90.78 / 64.48 & 90.78 / 64.48 \\
\hline
\hline
\multicolumn{6}{|l|}{\textbf{Distinct Info.~(Source / Target)}} \\
\hline
\multicolumn{1}{|c}{\multirow{6}{*}{SBS}}
& \multicolumn{1}{|l|}{Segment Num.} & \multicolumn{4}{c|}{4,746 / 57,032} \\
& \multicolumn{1}{|l|}{Avg Seg Len.} & 105.17 / 58.55 & 111.27 / 64.49 & 90.78 / 64.48 & 90.78 / 64.48 \\
\cline{2-6} 
& \multicolumn{1}{|l|}{MP PPROC: \makecell[l]{Time \\ Memory }} & \makecell[c]{131.33s \\ 4455.53 MB. } & \makecell[c]{80.42s \\ 7267.58 MB. } &  \makecell[c]{640.22s \\ 57924.36 MB. } &  \makecell[c]{133.01s \\ 7036.57 MB. } \\
\cdashline{3-6}
& \multicolumn{1}{|l|}{TK PPROC: \makecell[l]{Time \\ Memory }} & \makecell[c]{206.19s \\ 4478.97 MB. } & \makecell[c]{164.89s \\ 7291.22 MB. } & \makecell[c]{745.57s \\ 57948.21 MB. } & \makecell[c]{247.22s \\ 7052.71 MB. } \\
\hline
\multicolumn{1}{|c}{\multirow{6}{*}{ \makecell[c]{Blob \\ (Max~64) }}}
& \multicolumn{1}{|l|}{Segment Num.} & 4,083 / 38,828 & 4,189 / 40,971 & 3,752 / 41,706 & 3,752 / 41,706 \\
& \multicolumn{1}{|l|}{Avg Seg Len.} & 122.24 / 86.01 & 126.06 / 89.76 & 114.83 / 88,17 & 114.83 / 88,17 \\
\cline{2-6} 
& \multicolumn{1}{|l|}{MP PPROC: \makecell[l]{Time \\ Memory }} & \makecell[c]{107.54s \\ 4392.51 MB. } & \makecell[c]{70.92s \\ 7213.87 MB. }  & \makecell[c]{564.88s \\ 55890.82 MB. } & \makecell[c]{127.97s \\ 7023.15 MB. } \\
\cdashline{3-6}
& \multicolumn{1}{|l|}{TK PPROC: \makecell[l]{Time \\ Memory }} & \makecell[c]{164.87s \\ 4416.13 MB. } & \makecell[c]{139.41s \\ 7238.25 MB. } & \makecell[c]{655.54s \\ 55914.12 MB. }  & \makecell[c]{220.72s \\ 7040.32 MB. }  \\
\hline
\multicolumn{1}{|c}{\multirow{6}{*}{ \makecell[c]{OFLS \\ (FL~30, \\ OR~0.5) }}}
& \multicolumn{1}{|l|}{Segment Num.} & 33,151 / 222,149 & 35,082 / 244,688 & 28,594 / 244,653 & 28,594 / 244,653 \\
& \multicolumn{1}{|l|}{Avg Seg Len.} & 29.95 / 29.97 & 29.95 / 29.97 & 29.95 / 29.97 & 29.95 / 29.97 \\
\cline{2-6} 
& \multicolumn{1}{|l|}{MP PPROC: \makecell[l]{Time \\ Memory }} & \makecell[c]{71.38s \\ 2758.95 MB. } & \makecell[c]{49.25s \\ 1685.84 MB. }  & \makecell[c]{119.36s \\ 2338.35 MB. } &  \makecell[c]{380.51s \\ 3203.90 MB. } \\
\cdashline{3-6}
& \multicolumn{1}{|l|}{TK PPROC: \makecell[l]{Time \\ Memory }} & \makecell[c]{569.54s \\ 2782.64 MB. } & \makecell[c]{591.48s \\ 1715.25 MB. } & \makecell[c]{650.14s \\ 2370.38 MB. } & \makecell[c]{912.74s \\ 3236.67 MB. } \\
\hline
\end{tabular}
\end{adjustbox}
\caption{
The statistical information regarding the preprocessing steps before document alignment, where
“PPROC” represents for preprocessing.}
\vspace{-0.4cm}
\label{tab:info}
\end{table*}

\footnotetext{
Because the English documents contain duplicates, the number of gold pairs exceeds that of the Japanese documents.}

\begin{table*}[ht]
\centering
\resizebox{\textwidth}{!}{
\begin{tabular}{|ll|c|c|c|c|c|}
\hline
\multicolumn{2}{|l|}{\multirow{3}{*}{Strategies \& Models}} & \multicolumn{5}{c|}{Embedding Models}     \\ 
\cline{3-7} 
&  
& \href{https://huggingface.co/setu4993/LaBSE}{(a)~LaBSE}
& \href{https://huggingface.co/sentence-transformers/distiluse-base-multilingual-cased-v2}{(b)\makecell[c]{distiluse-base-\\ multi-cased-v2 }} 
& \href{https://github.com/facebookresearch/LASER}{(c)~LASER-2}
& \href{https://huggingface.co/BAAI/bge-m3}{(d)\makecell[c]{BGE M3\\ (dense only) }}
& \href{https://huggingface.co/jinaai/jina-embeddings-v3}{(e)~jina-embeddings-v3}   \\ 
\hline
\multicolumn{7}{|l|}{\textbf{Experiments}~\small(F1 Score \(\uparrow\) / PPROC. Time~(sec.) \(\downarrow\))} \\
\hline
\multicolumn{1}{|c}{\multirow{4}{*}{SBS}}
& \multicolumn{1}{|l|}{Mean-Pool}    & 0.8362 / 131.27s & 0.8362 / 80.40s  & 0.5862 / 543.10s & 0.8448 / 637.01s   & 0.8362 / 133.72s  \\
& \multicolumn{1}{|l|}{TK-PERT}      & 0.8448 / 206.19s & 0.8147 / 164.89s & 0.5819 / \underline{\textbf{652.32s}}  & 0.8362 / 745.57s & 0.8706 / 247.22s  \\
& \multicolumn{1}{|l|}{OT w/Mean}    & 0.8448 / 131.58s & 0.8448 / 80.46s & \underline{\textbf{0.4784}} / 543.87s  & 0.8621 / 642.20s  & 0.8578 / 132.73s  \\
& \multicolumn{1}{|l|}{BiMax w/Mean} & 0.8922$^{\dag}$$^{\ddag}$ / 131.47s & 0.9052$^{\dag}$$^{\ddag}$ / 80.49s  & 0.7414$^{\dag}$$^{\ddag}$ / 543.61s & 0.9181$^{\dag}$$^{\ddag}$ / 640.27s  & \underline{\textbf{0.9310$^{\dag}$$^{\ddag}$}} / 134.52s  \\
\hline
\multicolumn{1}{|c}{\multirow{4}{*}{ \makecell[c]{Blob \\ (Max~64) }}}
& \multicolumn{1}{|l|}{Mean-Pool}    & 0.8621 / 107.02s & \underline{\textbf{0.8663}} / 70.80s  & \underline{\textbf{0.5948}} / \underline{\textbf{533.63s}} & \underline{\textbf{0.8750}} / 565.45s  & \underline{\textbf{0.8448}} / \underline{\textbf{127.75s}}  \\
& \multicolumn{1}{|l|}{TK-PERT}      & 0.8663 / \underline{\textbf{164.87s}} & 0.8491 / \underline{\textbf{139.41s}} & 0.5905 / \underline{\textbf{640.51s}} & 0.8534 / 655.54s  & 0.8578 / \underline{\textbf{220.72s}} \\
& \multicolumn{1}{|l|}{OT w/Mean}    & 0.8233 / 107.84s & 0.8405 / 70.46s & 0.4439 / \underline{\textbf{533.61s}} & 0.8362 / 564.84s & 0.8276 / \underline{\textbf{128.12s}} \\
& \multicolumn{1}{|l|}{BiMax w/Mean} & 0.9009$^{\dag}$$^{\ddag}$ / 106.65s & 0.9052$^{\dag}$$^{\ddag}$ / 71.16s & 0.7586$^{\dag}$$^{\ddag}$ / \underline{\textbf{533.08s}} & 0.9181$^{\dag}$$^{\ddag}$ / 564.76s & 0.9052$^{\dag}$$^{\ddag}$ / \underline{\textbf{127.32s}} \\
\hline
\multicolumn{1}{|c}{\multirow{4}{*}{ \makecell[c]{OFLS \\ (FL~30, OR~0.5) }}}
& \multicolumn{1}{|l|}{Mean-Pool}    & \underline{\textbf{0.8707}} /  \underline{\textbf{71.59s}} & 0.8233 / \underline{\textbf{49.23s}} & 0.5302 / 1246.64s & 0.8491 / \underline{\textbf{119.38s}} & 0.7716 / 380.98s  \\
& \multicolumn{1}{|l|}{TK-PERT}      & \underline{\textbf{0.9483}} / 569.54s & \underline{\textbf{0.8966}} / 591.48s & \underline{\textbf{0.8134}} / 1860.80s & \underline{\textbf{0.9224}} / \underline{\textbf{650.14s}} & \underline{\textbf{0.9310}} / 912.74s \\
& \multicolumn{1}{|l|}{OT w/Mean}    & \underline{\textbf{0.9569}} /  \underline{\textbf{71.33s}} & \underline{\textbf{0.9397}} / \underline{\textbf{49.10s}} & 0.4354 / 1223.61s & \underline{\textbf{0.8879}} / \underline{\textbf{119.36s}} & \underline{\textbf{0.8966}} / 379.59s \\
& \multicolumn{1}{|l|}{BiMax w/Mean} & \underline{\textbf{0.9612}} /  \underline{\textbf{71.14s}} & \underline{\textbf{0.9569$^{\dag}$}} / \underline{\textbf{49.32s}} & \underline{\textbf{0.7845$^{\ddag}$}} / 1225.91s & \underline{\textbf{0.9483$^{\ddag}$}} / \underline{\textbf{119.36s}} & 0.9267$^{\ddag}$ / 381.05s \\
\hline
\end{tabular}
}
\caption{
The results for comparing SBS, Blob, and OFLS under each embedding model on the Ja-En MnRN dataset,
where
“FL” represents for fixed-length,
“OR” represents for overlapping rate,
“Max” represents the token limitation of Blob.
For each model and the four document alignment methods,
we underline and bold the \underline{\textbf{result}} that achieves the higher F1 score or shorter preprocessing time under SBS, Blob, or OFLS.
For each segmentation strategy within each model, 
$^{\dag}$ is appended when BiMax demonstrates statistically significant superiority over both Mean-Pool and TK-PERT, 
and $^{\ddag}$ is used when it is significantly superior to OT.}
\vspace{-0.5cm}
\label{tab:ana_result}
\end{table*}

\section{Analysis of Document Alignment Performance on MnRN}
\label{sec:anal}
We use the Ja-En MnRN dataset
to conduct the analysis 
under various sentence embedding models, three segmentation strategies, SBS\footnote{
Sentence-based Segmentation~(SBS): split a document into sentences 
using delimiters such as line breaks or periods.}, 
Blob\footnote{
Blob: concatenate multiple consecutive sentences as a single unit until reaching a specified limitation.
}~\citep{finkelstein-etal-2024-introducing},
and OFLS\footnote{
Overlapping Fixed-Length Segmentation~(OFLS): split a document into segments through a fixed-length sliding window, with a proportion of overlap between adjacent segments.
}~\citep{wang-etal-2024-document}, 
and four document alignment methods,
focusing on three main points:
\( (1) \) which model is suitable for which segmentation strategy, 
\( (2) \) how do different document alignment methods perform under each model,
and \( (3) \) which combination of these three factors yields the best results.

The reasons for selecting embedding models and the model settings 
are recorded in Appendix~\ref{sec:model_select} and \ref{sec:model_set},
while the experimental setup and the details of the evaluation metrics
are described in Appendix~\ref{sec:exper_set}.

We present the statistical information for four models under various segmentation strategies in Table~\ref{tab:info}.
Since Mean-Pool, OT w/Mean, and BiMax w/Mean require the same preprocessing steps for document alignment,
which include segmentation, segment embedding, and mean vector generation, 
we only use Mean-Pool~(MP) as a representative.
In contrast, TK-PERT~(TK) incurs additional time for LIDF\footnote{
LIDF is used for scaling segments based on the inverse
of the (linear, rather than logarithmic) number of documents
that contain the given segment.}
and the modified PERT distribution compared to MP, 
resulting in a longer preprocessing time that is dependent on the number of segments. 

\subsection{Performance Comparison}
\label{sec:per_com}
\noindent\( (1) \) \textit{Which model is suitable for which segmentation strategy?}\\
As shown in Table~\ref{tab:ana_result},
we present the results of five models labeled (a)$\sim$(e).
More detailed results for additional models can be found in Table~\ref{tab:extra_result} of Appendix~\ref{sec:model_select}.
For models (a), (b), and (d),
OFLS demonstrates an improvement in the F1 score in most cases 
and a reduction in preprocessing time~(except for TK-PERT) compared to the other two segmentation strategies.
However, for the LASER-2 model, 
although the use of OFLS improves the accuracy of the TK-PERT and BiMax methods, 
its performance on Mean-Pool and OT remains poor.
Additionally, the preprocessing speed is obviously diminished, 
which may be attributed to the chain structure of LSTM,  
due to the rise in the total number of tokens 
resulting from overlapping segments in OFLS.

Specifically, the jina-embeddings-v3 model achieves a relatively high F1 score 
when using the SBS segmentation, 
with a comparable speed to LaBSE.
Although employing OFLS may further enhance accuracy, 
the preprocessing time for the jina-embeddings-v3 model
becomes longer, which may be caused by the use of RoPE~\cite{rope}
and FlashAttention 2~\cite{dao2024flashattention2} mechanisms.

\begin{figure}[ht]
    \centering
    \includegraphics[scale=0.29]{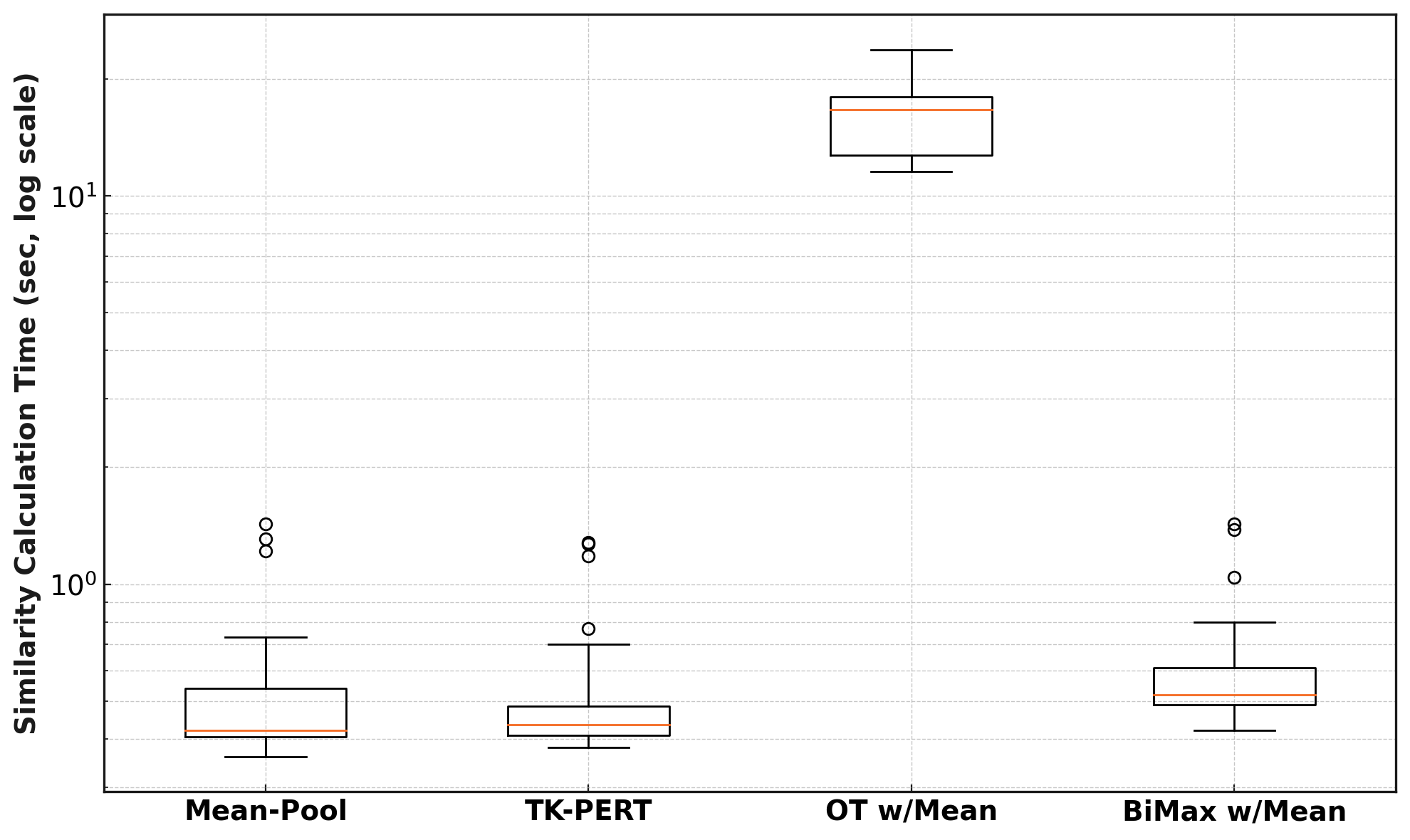}
    \caption{
    The similarity computation time~(in seconds, log scale) for the four document alignment methods.}
    \label{fig:boxplots_speed}
    \vspace{-0.3cm}
\end{figure}

Moreover, we provide an expanded discussion of Blob in Appendix~\ref{sec:blob_dis}.

\noindent\( (2) \) \textit{How do different document alignment methods perform under each model?}\\
Due to the limited scale of the MnRN dataset, 
the similarity computation times across different segmentation strategies 
and embedding models show minimal variation across the four document alignment methods.
Therefore, we present the distribution of these times in Figure~\ref{fig:boxplots_speed},
while Appendix~\ref{sec:model_select} provides detailed results.

Figure~\ref{fig:boxplots_speed} shows that the similarity computation time 
required for BiMax is much shorter than OT.
However, it should be noted that,
OT processes document pairs sequentially due to its optimization routine. 
In contrast, BiMax supports batched parallel computation.
For a fair runtime comparison, 
BiMax is limited to single-pair computation in this paper.
We follow~\citet{yeh-2000-accurate} and
conduct a statistical significance test~($p$ < $0.05$) 
between BiMax and the other three document alignment methods. 
The detailed process is described in Appendix~\ref{sec:exper_set}. 
Despite its lightweight design, 
BiMax outperforms competing methods in most scenarios 
and shows statistically significant gains in some cases.

\noindent\( (3) \) \textit{Which combination of these three factors yields the best results?}\\
Overall,
when using OFLS, 
LaBSE demonstrates superior accuracy compared to other models, and among the document alignment methods, 
according to Table~\ref{tab:ana_result},
BiMax achieves the best performance. 
The model closest to LaBSE under OFLS, distiluse-base-multilingual-cased-v2, 
while lower in accuracy, 
offers advantages in terms of speed and memory efficiency according to Table~\ref{tab:info}.

\section{Experiment on the WMT16 document alignment shared task}
\label{sec:wmt16_exp}
\vspace{0.15cm}
To test the BiMax method further, we conduct experiments on the WMT16 document alignment task. 
We use the same settings as \citet{wang-etal-2024-document} for a comparison with their work.
The detailed experimental setup and dataset information are recorded in Appendix~\ref{sec:exper_set}.

The results are presented in Table~\ref{tab:wmt16}.
Similarly, under the OFLS segmentation, 
the BiMax method improves recall by 0.3\% to 2.4\% compared to SBS. 
Compared with the results of \citet{wang-etal-2024-document}, 
the BiMax method demonstrates slightly higher accuracy than the OT and TK-PERT methods under SBS. 
However, the opposite trend is observed under OFLS.
Although BiMax cannot comprehensively outperform OT,
its speed achieves approximately a 100-fold increase relative to OT,
measured by the number of document pairs processed per second.
Rather than solely prioritizing precision,
this research emphasizes the efficiency of the method. 
Moreover, as noted in Section~\ref{sec:per_com}, 
BiMax can be executed in parallel via batch processing,
potentially resulting in faster similarity computations.

\vspace{-0.2cm}
\begin{table}[ht]
\centering
\small
\begin{adjustbox}{width=0.48\textwidth}
\begin{tabular}{lccccc}
\hline
Method & \makecell[c]{Segment \\ Strategy } & Recall\(\uparrow\) & \makecell[c]{Sim Speed \\ (pairs / sec.)}\(\uparrow\) \\
\hline
\multicolumn{4}{l}{\citet{wang-etal-2024-document}~(LaBSE)} \\
\hline
Mean-Pool              & SBS & 82.6\% & \multirow{4}{*}{97,358.98} \\
Mean-Pool              & OFLS & 92.6\% & \\
\cdashline{1-3}
TK-PERT                & SBS & 95.2\% & \\
TK-PERT                & OFLS & 96.3\% &  \\
\hdashline
OT w/Mean-Pool         & SBS & 90.6\% & \multirow{2}{*}{91.98} \\
OT w/TK-PERT           & SBS & 95.6\% & \\
\hdashline
OT w/Mean-Pool         & OFLS & \textbf{93.7\%} & \multirow{2}{*}{99.34} \\
OT w/TK-PERT           & OFLS & \textbf{96.8\%} & \\
\hline
\hline
\textbf{This work~(LaBSE)} \\
\hline
BiMax w/Mean-Pool         & SBS & \textbf{90.7\%} & \multirow{2}{*}{\textbf{11,510.92}} \\
BiMax w/TK-PERT           & SBS & \textbf{95.8\%} & \\
\hdashline
BiMax w/Mean-Pool         & OFLS & 93.1\% & \multirow{2}{*}{\textbf{13,220.15}} \\
BiMax w/TK-PERT           & OFLS & 96.1\% & \\
\hline
\end{tabular}
\end{adjustbox}
\caption{The results of soft recall on the WMT16 test data.
Between BiMax and OT, we highlight the superior result in \textbf{bold}.}
\vspace{-0.3cm}
\label{tab:wmt16}
\end{table}

\section{Experiments on Low-Resource Languages}
\label{sec:low_lang}

We use the dataset constructed by \citet{ref1},
which covers English, Sinhala, and Tamil 
(hereafter referred to as the Fernando dataset\footnote{
\url{https://github.com/kdissa/comparable-corpus}
})
to evaluate the effectiveness of BiMax.
The dataset comprises four web domains: Army, Hiru, ITN, and NewsFirst,
and three language pairs: En–Si, En–Ta, and Si–Ta. 
More detailed experimental settings and dataset statistics are provided in Appendix~\ref{sec:exper_set} and Appendix~\ref{sec:detail_low_rslt}.

We conduct experiments on three language pairs across four web domains.
For each language pair, the final recall is computed as the weighted average of its results over the domains, 
with weights determined by the number of gold pairs in each domain.
Detailed results are documented in Appendix~\ref{sec:detail_low_rslt}.

\begin{table}[ht]
\centering
\small
\begin{adjustbox}{width=0.48\textwidth}
\begin{tabular}{lcccc}
\hline
\multirow{2}{*}{Method} & \multirow{2}{*}{\makecell[c]{Segment \\ Strategy }} & \multicolumn{3}{c}{Language Pair} \\
\cline{3-5}
& & En-Si & En-Ta & Si-Ta \\
\hline
\multicolumn{5}{l}{LaBSE} \\
\hline
\multirow{2}{*}{Mean-Pool}     & SBS  & 93.10\% & 77.70\% & 79.41\% \\
                               & OFLS & 92.97\% & 86.39\% & 85.13\% \\
\cdashline{2-5}
\multirow{2}{*}{TK-PERT}       & SBS  & 89.32\% & 74.52\% & 75.48\% \\
                               & OFLS & 90.18\% & 82.64\% & 80.98\% \\
\cdashline{2-5}
\multirow{2}{*}{OT}& SBS  & 91.83\% & 78.94\% & 81.24\% \\
                               & OFLS & 92.51\% & 86.12\% & 86.74\% \\
\cdashline{2-5}
\multirow{2}{*}{BiMax} & SBS  & \textbf{95.53\%} & 83.85\% & 84.91\% \\
                                   & OFLS & 95.41\% & \textbf{91.33\%} & \textbf{89.71\%} \\
\hline
\end{tabular}
\end{adjustbox}
\caption{The results of recall on the Fernando dataset.
We highlight the best one in \textbf{bold} for each language pair.}
\vspace{-0.3cm}
\label{tab:low_lang_avg_recall}
\end{table}

As shown in Table~\ref{tab:low_lang_avg_recall}, 
BiMax outperforms the other three methods across all three language pairs, 
with a pronounced improvement observed in En–Ta.
Furthermore, in most cases, 
OFLS achieves better performance than SBS.
Even in the case of En–Si, where OFLS is slightly inferior to SBS,
the difference remains marginal.

Meanwhile, as a multi-way dataset, 
the same method exhibits considerable performance variation across different language pairs. 
In particular, when Tamil is used as the target language for retrieval, 
the accuracy differs substantially compared with En–Si, 
a discrepancy that may be attributed to variations in the embedding precision of the model across different languages.
Moreover, considering the dataset-specific characteristics,
factors beyond language, such as document length, 
may also affect alignment accuracy. 
Specifically, TK-PERT and OT are possibly better suited for handling long texts but perform less effectively on short texts. 
A more detailed analysis is provided in Appendix~\ref{sec:ana_doclen}.

\section{Conclusion}
\vspace{-0.1cm}
\label{sec:con}
This paper introduces a novel and efficient BiMax Score for the document alignment task,
reducing computational complexity compared to OT.
However, while BiMax shows the best performance on the
Fernando dataset and the small-scale MnRN dataset,
results from the WMT16 document alignment task reveal that 
we cannot definitively assert BiMax's accuracy surpasses OT or TK-PERT.
Instead, we advocate for BiMax primarily for its efficiency in scenarios such as 
processing large-scale web-crawled data. 
In these cases, according to our analysis of experiments, 
the LaBSE + OFLS + BiMax approach is recommended,
as it outperforms all other combinations.

\section{Limitations}

The existing publicly available datasets for document alignment are limited.
Even large-scale multilingual parallel document corpora such as CC-Aligned\footnote{
\url{https://www.statmt.org/cc-aligned/}}~\citep{el-kishky-etal-2020-ccaligned}, 
which consist of web pages aligned through automated document alignment methods, 
cannot guaranty ground truth due to the absence of manual verification.
In addition,
although we have explored the effectiveness of BiMax on low-resource languages,
the Fernando dataset~\citep{ref1} covers only Sinhala and Tamil. 
Since low-resource languages differ considerably from one another,
it cannot be guaranteed that the method generalizes equally well to all such languages.
Moreover, many other low-resource languages still lack established datasets.

Furthermore, although we evaluated multiple embedding models on the Ja-En MnRN dataset, 
the representational capabilities of different embedding models vary across languages.
Therefore, the LaBSE model may not consistently achieve optimal performance in all scenarios.

Finally, as discussed in Section~\ref{sec:wmt16_exp}, Section~\ref{sec:low_lang} and Section~\ref{sec:con}, 
its performance under OFLS does not surpass TK-PERT and OT on the WMT16 document alignment task.
Thus, we emphasize efficiency rather than solely pursuing precision.

\section{Ethical statement}

The embedding models used in this paper, 
LaBSE~\citep{feng-etal-2022-language}, 
LASER-2~\citep{heffernan-etal-2022-bitext}, 
LEALLA~\citep{mao-nakagawa-2023-lealla},
paraphrase-multilingual-MiniLM-L12-v2,
and distiluse-base-multilingual-cased-v2,
paraphrase-multilingual-mpnet-base-v2~\citep{reimers-gurevych-2019-sentence}, 
BGE M3~\citep{chen-etal-2024-m3}, 
and jina-embeddings-v3~\citep{sturua2024jinaembeddingsv3multilingualembeddingstask},
are publicly available for research. 

The WMT16 test data is provided by the WMT16 document alignment shared task~\citep{buck-koehn-2016-findings},
and the Fernando dataset has been publicly released by \citet{ref1}.
\bibliography{custom, anthology}

\appendix

\section{Embedding Model Selection}
\label{sec:model_select}
In Section~\ref{sec:anal},
first, we choose the LaBSE~\cite{feng-etal-2022-language} 
and LASER-2 models~\citep{heffernan-etal-2022-bitext},
which are frequently used for the bitext mining task,
and also include a knowledge-distilled, lightweight variant of LaBSE, 
the LEALLA model~\cite{mao-nakagawa-2023-lealla}.

Subsequently, we employ three representative multilingual 
models from the Sentence Transformers library\footnote{\url{https://huggingface.co/sentence-transformers}}:
paraphrase-multilingual-MiniLM-L12-v2,
distiluse-base-multilingual-cased-v2,
and paraphrase-multilingual-mpnet-base-v2~\citep{reimers-gurevych-2019-sentence}, 
which perform strongly on the STS task.

Finally, considering the MTEB benchmark~\cite{muennighoff-etal-2023-mteb},
which encompasses several embedding tasks,
we select two models that currently achieve state-of-the-art performance 
on the leaderboard\footnote{\url{https://huggingface.co/spaces/mteb/leaderboard}},
which are capable of processing long sentences
and suitable for multi-task scenarios: 
BGE M3~\cite{chen-etal-2024-m3}, 
and jina-embeddings-v3~\cite{sturua2024jinaembeddingsv3multilingualembeddingstask}.
Additionally,
we also consider the multi-e5-large model~\cite{wang2024multilingual}
and the paraphrase-multilingual-mpnet-base-v2 model~\cite{reimers-gurevych-2019-sentence}.
The results are presented in Table~\ref{tab:extra_result}.

\begin{sidewaystable*}
\centering
\resizebox{\textwidth}{!}{
\begin{tabular}{|ll|ccccccccc|}
\hline
\multicolumn{2}{|l|}{Strategies \& Models}
& \href{https://huggingface.co/setu4993/LaBSE}{(a)~LaBSE}
& \href{https://huggingface.co/setu4993/LEALLA-large}{(b)~LEALLA-large} 
& \href{https://huggingface.co/sentence-transformers/paraphrase-multilingual-MiniLM-L12-v2}{(c)~\makecell[c]{ paraphrase-multi- \\ MiniLM-L12-v2 }} 
& \href{https://huggingface.co/sentence-transformers/distiluse-base-multilingual-cased-v2}{(d)~\makecell[c]{distiluse-base-\\ multi-cased-v2 }} 
& \href{https://huggingface.co/sentence-transformers/paraphrase-multilingual-mpnet-base-v2}{(e)~\makecell[c]{paraphrase-multi- \\ mpnet-base-v2 }} 
& \href{https://github.com/facebookresearch/LASER}{(f)~LASER-2}
& \href{https://huggingface.co/intfloat/multilingual-e5-large}{(g)~multi-e5-large}
& \href{https://huggingface.co/BAAI/bge-m3}{(h)~BGE M3}
& \href{https://huggingface.co/jinaai/jina-embeddings-v3}{(i)~jina-embed-v3}   \\ 
\hline
\multicolumn{11}{|l|}{\textbf{Common Info.~(Source / Target)}} \\
\hline
\multicolumn{2}{|l}{Document Num.} & \multicolumn{9}{c|}{232 / 931} \\
\multicolumn{2}{|l}{Total Sentence Num.} & \multicolumn{9}{c|}{4,746 / 57,032} \\
\multicolumn{2}{|l}{Gold Pairs} & \multicolumn{9}{c|}{263} \\
\hline
\multicolumn{2}{|l}{Total Document Token Num.} & 0.50M / 3.34M & 0.50M / 3.34M & 0.43M / 3.68M & 0.53M / 3.68M & 0.43M / 3.68M & 0.57M / 4.47M & 0.43M / 3.68M & 0.43M / 3.68M & 0.43M / 3.68M \\
\multicolumn{2}{|l}{Average Sentence Token Num.} & 105.17 / 58.55 & 105.17 / 58.55 & 90.78 / 64.48 & 111.27 / 64.49 & 90.78 / 64.48 & 119.68 / 78.31 & 90.78 / 64.48 & 90.78 / 64.48 & 90.78 / 64.48 \\
\hline
\hline
\multicolumn{11}{|l|}{\textbf{Distinct Info.~(Source / Target)}} \\
\hline
\multicolumn{1}{|c}{\multirow{6}{*}{SBS}}
& \multicolumn{1}{|l|}{Segment Num.} & \multicolumn{9}{c|}{4,746 / 57,032} \\
& \multicolumn{1}{|l|}{Average Segment Len.} & 105.17 / 58.55 & 105.17 / 58.55 & 90.78 / 64.48 & 111.27 / 64.49 & 90.78 / 64.48 & 119.68 / 78.31 & 90.78 / 64.48 & 90.78 / 64.48 & 90.78 / 64.48 \\
\cline{2-11} 
& \multicolumn{1}{|l|}{MP PPROC: \makecell[l]{Time \\ Memory }} & \makecell[c]{131.33s \\ 4455.53 MB.} & \makecell[c]{60.78s \\ 1555.01 MB. } & \makecell[c]{59.01s \\ 1855.53 MB. } & \makecell[c]{80.42s \\ 7267.58 MB. } & \makecell[c]{148.88s \\ 5894.61 MB. } & \makecell[c]{543.20s \\ 3358.72 MB.} & \makecell[c]{458.74s \\ 8561.57 MB.}  & \makecell[c]{640.22s \\ 57924.36 MB. } &  \makecell[c]{133.01s \\ 7036.57 MB. } \\
\cdashline{3-11}
& \multicolumn{1}{|l|}{TK PPROC: \makecell[l]{Time \\ Memory }} & \makecell[c]{206.19s \\ 4478.97 MB. } & \makecell[c]{158.54s \\ 1562.89 MB. } & \makecell[c]{158.38s \\ 1867.35 MB. } & \makecell[c]{164.89s \\ 7291.22 MB. } & \makecell[c]{223.87s \\ 5918.26 MB. } & \makecell[c]{652.32s \\ 3406.95 MB. } & \makecell[c]{517.99s \\ 8593.10 MB. } & \makecell[c]{745.57s \\ 57948.21 MB. } & \makecell[c]{247.22s \\ 7052.71 MB. } \\
\hline
\multicolumn{1}{|c}{\multirow{6}{*}{ \makecell[c]{Blob \\ (Max~64) }}}
& \multicolumn{1}{|l|}{Segment Num.} & 4,083 / 38,828 & 4,083 / 38,828 & 3,752 / 41,706 & 4,189 / 40,971 & 3,752 / 41,706  & 4,198 / 46,761 & 3,752 / 41,706 & 3,752 / 41,706 & 3,752 / 41,706 \\
& \multicolumn{1}{|l|}{Average Segment Len.} & 122.24 / 86.01 & 122.24 / 86.01 & 114.83 / 88.17 & 126.06 / 89.76 & 114.83 / 88.17 & 135.30 / 95.51 & 114.83 / 88,17 & 114.83 / 88,17 & 114.83 / 88,17 \\
\cline{2-11} 
& \multicolumn{1}{|l|}{MP PPROC: \makecell[l]{Time \\ Memory }} & \makecell[c]{107.54s \\ 4392.51 MB. } & \makecell[c]{55.79s \\ 1535.52 MB. } & \makecell[c]{59.04s \\ 1832.44 MB. } & \makecell[c]{70.92s \\ 7213.87 MB. } & \makecell[c]{125.11s \\ 5840.50 MB. } & \makecell[c]{533.54s \\ 3343.34 MB. } & \makecell[c]{371.31s \\ 8495.56 MB. }  & \makecell[c]{564.88s \\ 55890.82 MB. } & \makecell[c]{127.97s \\ 7023.15 MB. } \\
\cdashline{3-11}
& \multicolumn{1}{|l|}{TK PPROC: \makecell[l]{Time \\ Memory }} & \makecell[c]{164.87s \\ 4416.13 MB. } & \makecell[c]{66.17s \\ 1543.25 MB. } & \makecell[c]{138.72s \\ 1844.26 MB. } & \makecell[c]{139.41s \\ 7238.25 MB. } & \makecell[c]{173.15s \\ 5864.29 MB. } & \makecell[c]{640.54s \\ 3391.38 MB. } & \makecell[c]{429.90s \\ 8527.09 MB. } & \makecell[c]{655.54s \\ 55914.12 MB. }  & \makecell[c]{220.72s \\ 7040.32 MB. }  \\
\hline
\multicolumn{1}{|c}{\multirow{6}{*}{ \makecell[c]{OFLS \\ (FL~30, OR~0.5) }}}
& \multicolumn{1}{|l|}{Segment Num.} & 33,151 / 222,149 & 33,151 / 222,149 & 28,594 / 244,653 & 35,082 / 244,688 & 28,594 / 244,653 & 37,742 / 297,245 & 28,594 / 244,653 & 28,594 / 244,653 & 28,594 / 244,653 \\
& \multicolumn{1}{|l|}{Average Segment Len.} & 29.95 / 29.97 & 29.95 / 29.97 & 29.95 / 29.97 & 29.95 / 29.97 & 29.95 / 29.97 & 29.96 / 29.98 & 29.95 / 29.97 & 29.95 / 29.97 & 29.95 / 29.97 \\
\cline{2-11} 
& \multicolumn{1}{|l|}{MP PPROC: \makecell[l]{Time \\ Memory }} & \makecell[c]{71.38s \\ 2758.95 MB.} & \makecell[c]{52.60s \\ 900.48 MB.} & \makecell[c]{49.06s \\ 966.69 MB. } & \makecell[c]{49.25s \\ 1685.84 MB. } & \makecell[c]{74.58s \\ 2070.51 MB. } & \makecell[c]{1221.74s \\ 1871.28 MB. } & \makecell[c]{259.11s \\ 3476.11 MB. } & \makecell[c]{119.36s \\ 2338.35 MB. } &  \makecell[c]{380.51s \\ 3203.90 MB. } \\
\cdashline{3-11}
& \multicolumn{1}{|l|}{TK PPROC: \makecell[l]{Time \\ Memory }} & \makecell[c]{569.54s \\ 2782.64 MB. } & \makecell[c]{548.93s \\ 908.37 MB. } & \makecell[c]{578.17s \\ 978.53 MB. } & \makecell[c]{591.48s \\ 1715.25 MB. } & \makecell[c]{599.66s \\ 2094.20 MB. } & \makecell[c]{1860.80s \\ 1920.33 MB. } & \makecell[c]{745.20s \\ 3507.69 MB. } & \makecell[c]{650.14s \\ 2370.38 MB. } & \makecell[c]{912.74s \\ 3236.67 MB. } \\
\hline
\end{tabular}
}
\caption{
The statistical information regarding the preprocessing steps before document alignment across various models and segmentation strategies,
where
“MP” represents for Mean-Pool,
“TK” represents for TK-PERT,
“PPROC” represents for preprocessing.}
\label{tab:info_all}
\end{sidewaystable*}

\begin{sidewaystable*}
\centering
\resizebox{\textwidth}{!}{
\begin{tabular}{|ll|c|c|c|c|c|c|c|c|c|}
\hline
\multicolumn{2}{|l|}{\multirow{3}{*}{Info. \& Methods}} & \multicolumn{8}{c|}{Embedding Models}     \\ 
\cline{3-11} 
&  
& \href{https://huggingface.co/setu4993/LaBSE}{(a)~LaBSE}
& \href{https://huggingface.co/setu4993/LEALLA-large}{(b)~LEALLA-large} 
& \href{https://huggingface.co/sentence-transformers/paraphrase-multilingual-MiniLM-L12-v2}{(c)~\makecell[c]{ paraphrase-multi- \\ MiniLM-L12-v2 }} 
& \href{https://huggingface.co/sentence-transformers/distiluse-base-multilingual-cased-v2}{(d)~\makecell[c]{distiluse-base-\\ multi-cased-v2 }} 
& \href{https://huggingface.co/sentence-transformers/paraphrase-multilingual-mpnet-base-v2}{(e)~\makecell[c]{paraphrase-multi- \\ mpnet-base-v2 }} 
& \href{https://github.com/facebookresearch/LASER}{(f)~LASER-2}
& \href{https://huggingface.co/intfloat/multilingual-e5-large}{(g)~multi-e5-large}
& \href{https://huggingface.co/BAAI/bge-m3}{(h)\makecell[c]{BGE M3\\ (dense only) }}
& \href{https://huggingface.co/jinaai/jina-embeddings-v3}{(i)~jina-embeddings-v3}   \\ 
\hline
\multicolumn{11}{|l|}{\textbf{Model Info.}} \\
\hline    
\multicolumn{2}{|l|}{Suitable Task} 
& Bitext.     & Bitext.     & STS         & STS         & STS         & Bitext.     & Multi-task  & Multi-task  & Multi-task  \\
\multicolumn{2}{|l|}{\#Param.}      
& 471M        & 147M        & 118M        & 135M        & 278M        & 43M         & 560M        & 567M        & 572M        \\
\multicolumn{2}{|l|}{\#Dim.}        
& 768         & 256         & 384         & 512         & 768         & 1024        & 1024        & 1024        & 1024        \\
\multicolumn{2}{|l|}{\#Lang.}       
& Multi.      & Multi.      & Multi.      & Multi.      & Multi.      & Mono.       & Multi.      & Multi.      & Multi.      \\
\multicolumn{2}{|l|}{\#Arch.}       
& Transformer & Transformer & Transformer & Transformer & Transformer & LSTM        & Transformer & Transformer & Transformer \\
\hline
\hline
\multicolumn{11}{|l|}{\textbf{Experiments}~\small(F1 Score \(\uparrow\) / PPROC. Time~(sec.) \(\downarrow\) / Sim. Time~(sec.) \(\downarrow\))} \\
\hline
\multicolumn{1}{|c}{\multirow{4}{*}{SBS}}
& \multicolumn{1}{|l|}{Mean-Pool}    & 0.8362 / 131.27s / 0.42s  & 0.3750 / 60.54s / 0.37s & 0.7543 / 59.00s / 0.36s   & 0.8362 / 80.40s  / 0.40s & 0.7716 / 148.60s / 0.46s  & 0.5862 / 543.10s / 0.42s  & 0.7802 / 457.94s / 0.42s  & 0.8448 / 637.01s / 1.43s  & 0.8362 / 133.72s / 0.68s \\
& \multicolumn{1}{|l|}{TK-PERT}      & 0.8448 / 206.19s / 0.48s  & \textbf{0.5129} / 158.54s / 0.39s & 0.7845 / 158.38s / 0.47s  & 0.8147 / 164.89s / 0.42s & 0.7931 / 223.87s / 0.41s  & 0.5819 / 652.32s / 0.41s  & 0.7845 / 517.99s / 0.46s  & 0.8362 / 745.57s / 1.27s  & 0.8706 / 247.22s / 0.77s \\
& \multicolumn{1}{|l|}{OT w/Mean}    & 0.8448 / 131.58s / 22.87s & 0.4525 / 60.87s  / 17.26s& 0.7845 / 58.98s  / 16.69s & 0.8448 / 80.46s  / 20.17s& 0.7974 / 149.07s / 18.13s & 0.4784 / 543.87s / 15.54s & 0.8060 / 461.78s / 15.43s & 0.8621 / 642.20s / 17.62s & 0.8578 / 132.73s / 17.90s \\
& \multicolumn{1}{|l|}{BiMax w/Mean} & \textbf{0.8922$^{\dag}$$^{\ddag}$} / 131.47s / 0.49s  & 0.4655 / 60.83s  / 0.43s & \textbf{0.8319$^{\dag}$$^{\ddag}$} / 59.35s  / 0.42s  & \textbf{0.9052$^{\dag}$$^{\ddag}$} / 80.49s  / 0.46s & \textbf{0.8577$^{\dag}$$^{\ddag}$} / 148.40s / 0.49s  & \textbf{0.7414$^{\dag}$$^{\ddag}$} / 543.61s / 0.50s  & \textbf{0.8750$^{\dag}$$^{\ddag}$} / 462.17s / 0.80s  & \textbf{0.9181$^{\dag}$$^{\ddag}$} / 640.27s / 0.53s  & \textbf{0.9310$^{\dag}$$^{\ddag}$} / 134.52s / 0.70s  \\
\hline
\multicolumn{1}{|c}{\multirow{4}{*}{ \makecell[c]{Blob \\ (Max~64) }}}
& \multicolumn{1}{|l|}{Mean-Pool}    & 0.8621 / 107.02s / 0.42s & 0.3491 / 55.79s / 0.38s & 0.7672 / 59.04s / 0.36s & 0.8663 / 70.80s / 0.41s  & 0.7802 / 125.11s / 0.41s & 0.5948 / 533.63s / 0.41s & 0.7844 / 370.11s / 0.71s & 0.8750 / 565.45s / 1.31s & 0.8448 / 127.75s / 0.62s  \\
& \multicolumn{1}{|l|}{TK-PERT}      & 0.8663 / 164.87s / 0.45s  & \textbf{0.5129} / 119.17s / 0.38s & 0.7155 / 138.72s / 0.40s & 0.8491 / 139.41s / 0.39s & 0.7241 / 173.15 / 0.41s & 0.5905 / 640.51s / 0.42s & 0.7672 / 429.90s / 0.50s & 0.8534 / 655.54s / 1.28s & 0.8578 / 220.72s / 0.70s \\
& \multicolumn{1}{|l|}{OT w/Mean}    & 0.8233 / 107.84s / 23.83s & 0.4525 / 56.20s / 16.93s & 0.7802 / 59.14s / 18.38s & 0.8405 / 70.46s / 20.91s & 0.7802 / 125.23s / 18.69s & 0.4439 / 533.61s / 14.99s & 0.7974 / 371.94s / 16.07s & 0.8362 / 564.84s / 17.92s & 0.8276 / 128.12s / 17.10s \\
& \multicolumn{1}{|l|}{BiMax w/Mean} & \textbf{0.9009$^{\dag}$$^{\ddag}$} / 106.65s / 0.50s & 0.4699 / 56.29s / 0.49s & \textbf{0.8147$^{\dag}$$^{\ddag}$} / 59.16s / 0.43s & \textbf{0.9052$^{\dag}$$^{\ddag}$} / 71.16s / 0.52s & \textbf{0.8534$^{\dag}$$^{\ddag}$} / 125.21s / 0.43s & \textbf{0.7586$^{\dag}$$^{\ddag}$} / 533.08s / 0.51s & \textbf{0.8707$^{\dag}$$^{\ddag}$} / 371.15s / 0.64s & \textbf{0.9181$^{\dag}$$^{\ddag}$} / 564.76s / 1.43s & \textbf{0.9052$^{\dag}$$^{\ddag}$} / 127.32s / 0.68s \\
\hline
\multicolumn{1}{|c}{\multirow{4}{*}{ \makecell[c]{OFLS \\ (FL~30, OR~0.5) }}}
& \multicolumn{1}{|l|}{Mean-Pool}    & 0.8707 /  71.59s / 0.42s  & 0.3836 / 52.76s  / 0.44s & 0.7759 / 49.06s  / 0.36s  & 0.8233 / 49.23s / 0.43s & 0.7112 / 74.56s  / 0.40s   & 0.5302 / 1226.64s / 0.41s  & 0.7543 / 259.61s / 0.44s  & 0.8491 / 119.38s / 1.22s  & 0.7716 / 380.98s / 0.73s \\
& \multicolumn{1}{|l|}{TK-PERT}      & 0.9483 / 569.54s / 0.40s  & \textbf{0.6034} / 548.93s / 0.40s & 0.8707 / 578.17s / 0.48s & 0.8966 / 591.48s / 0.42s& 0.8793 / 599.66s / 0.40s   & \textbf{0.8134} / 1860.80s / 0.42s  & 0.8534 / 745.20s / 0.45s  & 0.9224 / 650.14s / 1.18s  & \textbf{0.9310} / 912.74s / 0.63s \\
& \multicolumn{1}{|l|}{OT w/Mean}    & 0.9569 /  71.33s / 12.67s & 0.4782 / 52.47s  / 12.67s& 0.8578 / 49.08s  / 11.64s & 0.9397 / 49.10s / 12.47s& 0.8922 / 74.31s  / 11.54s  & 0.4354 / 1223.61s / 12.78s & 0.7801 / 258.70s / 12.21s & 0.8879 / 119.36s / 12.97s & 0.8966 / 379.59s /12.44s \\
& \multicolumn{1}{|l|}{BiMax w/Mean} & \textbf{0.9612} /  71.14s / 0.49s  & 0.5348$^{\ddag}$ / 52.93s  / 0.53s & \textbf{0.9052$^{\dag}$$^{\ddag}$} / 49.09s  / 0.51s  & \textbf{0.9569$^{\dag}$} / 49.32s / 0.55s & \textbf{0.9138$^{\dag}$} / 74.47s  / 0.53s   & 0.7845$^{\ddag}$ / 1205.91s / 0.54s  & \textbf{0.9181$^{\dag}$$^{\ddag}$} / 258.35s / 0.58s  & \textbf{0.9483$^{\ddag}$} / 119.36s / 1.38s  & 0.9267$^{\ddag}$ / 381.05s / 1.04s \\
\hline
\end{tabular}
}
\caption{
The results from various sentence embedding models, segmentation strategies, 
and document alignment methods on the MnRN dataset.
where
“\#Param.” represents for the number of parameters,
“\#Dim.” represents for the embedding dimension,
“\#Lang.” represents for the language mode~(multilingual or monolingual),
“\#Arch.” represents for the model architecture,
“PPROC Time” represents for preprocessing time,
“SBS” represents for sentence-based segmentation,
“OFLS” represents for overlapping fixed-length segmentation,
“FL” represents for fixed-length,
“OR” represents for overlapping rate,
“Max” represents the token limitation of Blob.
Moreover, we put the highest F1 scores achieved by each model 
under each segmentation strategy in \textbf{bold}. 
For each segmentation strategy within each model, 
$^{\dag}$ is appended when BiMax demonstrates statistically significant superiority over both Mean-Pool and TK-PERT, 
and $^{\ddag}$ is used when it is significantly superior to OT.}
\label{tab:extra_result}
\vspace{-0.5cm}
\end{sidewaystable*}

\section{Embedding Model Settings}
\label{sec:model_set}
We maintain the default configurations for all models,
as these configurations represent the most general use cases. 
However, 
BGE M3 employs a half-precision floating-point format (fp16) by default, 
whereas most other models utilize a single-precision floating-point format (fp32). 
Furthermore, BGE M3 and LASER-2 generate vectors in the form of NumPy arrays, 
while other models predominantly output tensors or offer tensor output as an option. 
To establish method consistency, 
we implement a standardization protocol, converting all vectors to fp32 format 
and utilizing tensors after the embedding process.

Meanwhile, given that all models except LASER-2 
are derived from Hugging Face\footnote{\url{https://huggingface.co/}} ,
we can achieve substantial uniformity in the Python library and code framework, 
thereby facilitating meaningful comparisons of inference speeds across models. 
However, due to the LASER-2 model's different library and code program, 
absolute parity in comparative speed analysis between LASER-2 and other models cannot be established.

Because of the multifunctionality of the three multi-task models,  
we specify distinct usage.  
For the multi-e5-large model,  
which can leverage a prefix~(either “query:” or “passage:”) as the start of the text,  
after testing with some combinations or omitting the prefix altogether,  
we find that appending “query:” to both the source and target produces the highest accuracy.  
Regarding the BGE M3 model, which provides three functions for generating different scores,  
we elect to use only its dense embedding as output.  
Finally, for the jina-embeddings-v3 model,  
which offers a selection among various
LoRA adapters~\cite{DBLP:journals/corr/abs-2106-09685} depending on the desired task,  
we choose the “text-matching” task.

\section{Experiment Settings}
\label{sec:exper_set}
We follow the experimental settings of \citet{thompson-koehn-2020-exploiting} and \citet{wang-etal-2024-document},
configuring the hyper-parameters for the WMT16 document alignment task 
and the MnRN dataset in the TK-PERT method as 
$J=16, \gamma=20$ and $J=8, \gamma=16$,
respectively. 
The setting of the Fernando dataset is the same as the WMT16 test data.
Here, $J$ determines the number of windows produced by the TK-PERT method, 
while $\gamma$ is a hyper-parameter that controls the peakedness of the modified PERT distribution.
For OT, GMD, and BiMax, we retrieve 20, 32, and 32 candidates for
each source document in the MnRN dataset, 
WMT test data, and Fernando dataset,
respectively.
We put the basic information of the three datasets in Table~\ref{tab:data_info1} and Table~\ref{tab:data_info2}\footnote{
We use the data provided by the authors on GitHub, 
whose size differs from that reported in the original paper}.
For the WMT16 test data, 
we configure OFLS with a sliding-window size of 100 and an overlap ratio of 0.5,
the same as \citet{wang-etal-2024-document}.
For the Fernando dataset, 
we use a fixed-length window of 30 with the same overlap ratio.
In our Faiss Search~\citep{johnson2019billion} setup, we use IndexFlatIP as the index type and perform cosine similarity searches on GPUs.

\begin{table}[ht]
\centering
\begin{adjustbox}{width=0.47\textwidth}
\begin{tabular}{lcc}
\hline
                    & WMT16 test data & MnRN dataset \\
\hline
En Docs.            & 682k   & 931        \\
Fr Docs.            & 522k   & -          \\
Ja Docs.            & -      & 232        \\
Web domains         & 203    & 4          \\
Gold Pairs          & 2,402  & 263        \\
Direction           & Fr-En  & Ja-En      \\
\hline
\end{tabular}
\end{adjustbox}
\caption{Basic information for the WMT16 test data and MnRN dataset.}
\label{tab:data_info1}
\vspace{-0.6cm}
\end{table}

\begin{table}[ht]
\centering
\begin{adjustbox}{width=0.47\textwidth}
\begin{tabular}{llllllll}
\hline
\multirow{2}{*}{Web-domain} & \multicolumn{3}{c}{No. of Docs.}& & \multicolumn{3}{c}{Aligned Docs.} \\
\cline{2-4} \cline{6-8}
           & En    & Si    & Ta        &       & En-Si & En-Ta & Si-Ta             \\
\hline
Army       & 2,081 & 2,033 & 1,905     &       & 1,848 & 1,671 & 1,578             \\
Hiru       & 1,634 & 3,133 & 2,886     &       & 1,397 & 1,056 & 2,002             \\
ITN        & 1,942 & 4,898 & 1,521     &       & 352   & 112   & 34                \\
NewsFirst  & 2,278 & 1,821 & 2,333     &       & 344   & 316   & 97                \\
\hline
\end{tabular}
\end{adjustbox}
\caption{Basic information for the Fernando dataset.}
\label{tab:data_info2}
\end{table}

The final document alignment output follows a 1-1 rule~\citep{buck-koehn-2016-findings}, 
whereby each document ID should appear only once in the results. 
Consequently, we rank all matched document pairs by similarity and eliminate 
any lower-ranked pairs that contain a document ID already assigned at a higher rank.

For evaluation of the WMT16 document alignment shared task,
we adhere to previous 
work~\citep{buck-koehn-2016-findings, thompson-koehn-2020-exploiting,sannigrahi-etal-2023-best, wang-etal-2024-document}
via a ``soft'' recall metric, 
which assigns credit to document pairs 
if either the English or French document~(but not both) 
deviates from the reference document pair by less than 5\%, based on text edit distance. 
For the MnRN dataset, we follow~\citet{wang-etal-2024-document} in using the F1 score for evaluation. 
Since multiple correct target documents may correspond to a single source document in MnRN dataset,
both precision and recall are calculated with 
respect to the source-side instances within the set of gold pairs~(i.e., 
although there are 263 gold pairs, 
they involve only 232 unique source documents;
therefore, we define the total number of instances as 232.)

For significance testing, we refer to~\citet{yeh-2000-accurate} and adopt a randomization test procedure. 
Given two result sets, \( A \) and \( B \), 
with corresponding F1 scores \( \text{F1}_A \) 
and \( \text{F1}_B \)~(assuming \( \text{F1}_A > \text{F1}_B \)), 
we retain their intersection \( A \cap B \) 
and isolate the symmetric difference \( A \triangle B \), 
irrespective of the correctness of each pair.
The elements in \( A \triangle B \) are then randomly partitioned into two subsets, 
yielding \( 2^{|A \triangle B|} \) possible permutations. 
For each trial, the two subsets are combined with \( A \cap B \) 
to form new result sets \( A' \) and \( B' \), 
from which updated F1 scores \( \text{F1}_{A'} \) and \( \text{F1}_{B'} \) are computed. 
Let \( n \) denote the number of trials 
in which \( (\text{F1}_{A'} - \text{F1}_{B'} > \text{F1}_A - \text{F1}_B) \); 
the p-value is then calculated as \( (n + 1)/(2^{|A \triangle B|} + 1) \). 
Following~\citet{yeh-2000-accurate}, 
when \( |A \triangle B| > 20 \), we use an approximate randomization with 1,048,576 shuffles.

All experiments are conducted on two A6000 GPUs and one H100 GPU.

\section{Discussion of Blob and its Variants}
\label{sec:blob_dis}
We select four well-performing models in Section~\ref{sec:per_com}
to investigate the token limitation of Blob.
The results are presented in Table~\ref{tab:blob}.

As the token limitation increases,
the number of blobs segmented from the document decreases accordingly, 
resulting in a natural reduction in embedding computation time.
Furthermore, in most cases, 
the highest F1 Score is achieved with a token limitation of 64. 
Therefore, prioritizing accuracy, we compare the results in this case with SBS and OFLS in Section~\ref{sec:per_com}.

\begin{table}[ht]
\centering
\begin{adjustbox}{width=0.48\textwidth}
\begin{tabular}{|ll|c|c|c|c|}
\hline
\multicolumn{2}{|l|}{\multirow{3}{*}{Strategies \& Models}} & \multicolumn{4}{c|}{Embedding Models}     \\ 
\cline{3-6} 
&  
& \href{https://huggingface.co/setu4993/LaBSE}{LaBSE}
& \href{https://huggingface.co/sentence-transformers/distiluse-base-multilingual-cased-v2}{\makecell[c]{distiluse-base-\\ multi-cased-v2 }} 
& \href{https://huggingface.co/BAAI/bge-m3}{\makecell[c]{BGE M3\\ (dense only) }}
& \href{https://huggingface.co/jinaai/jina-embeddings-v3}{jina-embeddings-v3}   \\ 
\hline
\multicolumn{6}{|l|}{\textbf{Experiments}~\small(F1 Score \(\uparrow\) / PPROC. Time~(sec.) \(\downarrow\))} \\
\hline
\multicolumn{1}{|c}{\multirow{4}{*}{ \makecell[c]{Blob \\ (Max~64) }}}
& \multicolumn{1}{|l|}{Mean-Pool}    & 0.8621  /  107.02s  & \underline{\textbf{0.8663}} / 70.80s & 0.8750 / 565.45s & \underline{\textbf{0.8448}} / 127.75s  \\
& \multicolumn{1}{|l|}{TK-PERT}      & \underline{\textbf{0.8663}} / 164.87s & \underline{\textbf{0.8491}} / 139.41s & 0.8534 / 655.54 & 0.8578 / 220.72s \\
& \multicolumn{1}{|l|}{OT w/Mean}    & \underline{\textbf{0.8233}} / 107.84s & \underline{\textbf{0.8405}} / 70.46s & \underline{\textbf{0.8362}} / 564.84s & \underline{\textbf{0.8276}} / 128.12s \\
& \multicolumn{1}{|l|}{BiMax w/Mean} & 0.9009 /  106.65s  & \underline{\textbf{0.9052}} / 71.16s & 0.9181 / 564.76s & \underline{\textbf{0.9052}} / 127.32s \\
\hline
\multicolumn{1}{|c}{\multirow{4}{*}{ \makecell[c]{Blob \\ (Max~128) }}}
& \multicolumn{1}{|l|}{Mean-Pool}    & \underline{\textbf{0.8879}} / 75.65s & 0.8233 / 52.60s   & 0.8491 / 581.87s & \underline{\textbf{0.8448}} / 110.10s  \\
& \multicolumn{1}{|l|}{TK-PERT}      & 0.8621 / 109.81s  & \underline{\textbf{0.8491}} / 94.63s & 0.8664 / 640.70s & \underline{\textbf{0.8707}} / 169.36s \\
& \multicolumn{1}{|l|}{OT w/Mean}    & 0.8190 /  75.74s  & 0.8103 / 52.73s  & 0.8017 / 582.46s  & 0.8060 / 108.08s \\
& \multicolumn{1}{|l|}{BiMax w/Mean} & \underline{\textbf{0.9138}} /  75.98s & 0.8621 / 52.69s  & \underline{\textbf{0.9267}} / 581.77s & 0.9009 / 108.94s \\
\hline
\multicolumn{1}{|c}{\multirow{4}{*}{ \makecell[c]{Blob \\ (Max~256) }}}
& \multicolumn{1}{|l|}{Mean-Pool}    & 0.8793 / 52.09s & 0.8362 / 38.53s  & 0.8491 / 475.60s & 0.8147 / 99.09s  \\
& \multicolumn{1}{|l|}{TK-PERT}      & 0.8491 / 74.41s & 0.8405 / 64.89s  & 0.8578 / 510.18s & 0.8276 / 136.84s \\
& \multicolumn{1}{|l|}{OT w/Mean}    & 0.7543 / 51.96s & 0.7457 / 38.70s  & 0.7629 / 475.64s & 0.7457 / 99.26s \\
& \multicolumn{1}{|l|}{BiMax w/Mean} & 0.8879 / 51.97s & 0.8879 / 38.48s  & \underline{\textbf{0.9267}} / 475.76s & 0.8966 / 99.15s \\
\hline
\multicolumn{1}{|c}{\multirow{4}{*}{ \makecell[c]{Blob \\ (Max~384) }}}
& \multicolumn{1}{|l|}{Mean-Pool}    & 0.8706  /  \underline{\textbf{45.00s}}  & 0.8362 / \underline{\textbf{34.21s}} & \underline{\textbf{0.9138}} / \underline{\textbf{424.74s}} & 0.8190 / \underline{\textbf{95.54s}} \\
& \multicolumn{1}{|l|}{TK-PERT}      & 0.8147  /  \underline{\textbf{62.72s}}  & 0.7802 / \underline{\textbf{54.38s}} & \underline{\textbf{0.8879}} / \underline{\textbf{466.63s}} & 0.8621 / \underline{\textbf{125.52s}} \\
& \multicolumn{1}{|l|}{OT w/Mean}    & 0.6552  /  \underline{\textbf{44.97s}}  & 0.6336 / \underline{\textbf{34.41s}} & 0.6853 / \underline{\textbf{424.56s}} & 0.6638 / \underline{\textbf{95.42s}} \\
& \multicolumn{1}{|l|}{BiMax w/Mean} & 0.8793  /  \underline{\textbf{45.09s}}  & 0.8232 / \underline{\textbf{34.30s}} & 0.8879 / \underline{\textbf{424.34s}} & 0.8966 / \underline{\textbf{95.79s}} \\
\hline
\end{tabular}
\end{adjustbox}
\caption{
The results of different max token limitation settings for Blob. Each model's highest F1 scores and shortest preprocessing time are tagged in \textbf{\underline{bold}}.}
\label{tab:blob}
\vspace{-0.3cm}
\end{table}

\begin{table*}[ht]
\centering
\begin{adjustbox}{width=0.95\textwidth}
\begin{tabular}{|ll|c|c|c|c|}
\hline
\multicolumn{2}{|l|}{\multirow{3}{*}{Strategies \& Models}} & \multicolumn{4}{c|}{Embedding Models}     \\ 
\cline{3-6} 
&  
& \href{https://huggingface.co/setu4993/LaBSE}{(a)~LaBSE}
& \href{https://huggingface.co/sentence-transformers/distiluse-base-multilingual-cased-v2}{(b)\makecell[c]{distiluse-base-\\ multi-cased-v2 }} 
& \href{https://huggingface.co/BAAI/bge-m3}{(c)\makecell[c]{BGE M3\\ (dense only) }}
& \href{https://huggingface.co/jinaai/jina-embeddings-v3}{(d)~jina-embeddings-v3}   \\ 
\hline
\multicolumn{6}{|l|}{\textbf{Experiments of Blob-o w/tok}~\small(F1 Score \(\uparrow\) / PPROC. Time~(sec.) \(\downarrow\))} \\
\hline
\multicolumn{1}{|c}{\multirow{4}{*}{ \makecell[c]{Blob \\ (Max~64) }}}
& \multicolumn{1}{|l|}{Mean-Pool}    & 0.8621 / 107.02s & 0.8663 / 70.80s & \underline{\textbf{0.8750}} / \underline{\textbf{565.45s}} & 0.8448 / 127.75s  \\
& \multicolumn{1}{|l|}{TK-PERT}      & 0.8663 / 164.87s & 0.8491 / 139.41s & 0.8534 / \underline{\textbf{655.54s}} & 0.8578 / 220.72s \\
& \multicolumn{1}{|l|}{OT w/Mean}    & 0.8233 / 107.84s & \underline{\textbf{0.8405}} / 70.46s & \underline{\textbf{0.8362}} / \underline{\textbf{564.84s}} & \underline{\textbf{0.8276}} / 128.12s \\
& \multicolumn{1}{|l|}{BiMax w/Mean} & 0.9009 / 106.65s & \underline{\textbf{0.9052}} / 71.16s & 0.9181 / \underline{\textbf{564.76s}} & 0.9052 / 127.32s \\
\hline
\multicolumn{1}{|c}{\multirow{4}{*}{ \makecell[c]{Blob-o w/tok \\ (64, 0.15) }}}
& \multicolumn{1}{|l|}{Mean-Pool}    & \underline{\textbf{0.9224}} / 115.24s & \underline{\textbf{0.8707}} / 77.39s  & 0.8448 / 594.08s & \underline{\textbf{0.8534}} / 136.96s \\
& \multicolumn{1}{|l|}{TK-PERT}      & \underline{\textbf{0.8966}} / 178.46s & 0.8664 / 153.17s & 0.8664 / 686.60s & \underline{\textbf{0.8922}} / 231.78s \\
& \multicolumn{1}{|l|}{OT w/Mean}    & \underline{\textbf{0.8793}} / 115.38s & 0.8147 / 77.26s  & \underline{\textbf{0.8362}} / 593.45s & \underline{\textbf{0.8276}} / 137.20s \\
& \multicolumn{1}{|l|}{BiMax w/Mean} & 0.9181 / 114.95s & 0.8922 / 77.14s  & 0.9095 / 594.40s & \underline{\textbf{0.9267}} / 137.36s \\
\hline
\multicolumn{1}{|c}{\multirow{4}{*}{ \makecell[c]{Blob-o w/tok \\ (128, 0.15) }}}
& \multicolumn{1}{|l|}{Mean-Pool}    & 0.8966 / \underline{\textbf{82.35s}}  & 0.8491 / \underline{\textbf{59.84s}}  & 0.8707 / 599.33s & 0.8491 / \underline{\textbf{118.38s}} \\
& \multicolumn{1}{|l|}{TK-PERT}      & 0.8879 / \underline{\textbf{124.02s}} & \underline{\textbf{0.8707}} / \underline{\textbf{110.24s}} & \underline{\textbf{0.8750}} / 672.73s & 0.8578 / \underline{\textbf{182.58s}} \\
& \multicolumn{1}{|l|}{OT w/Mean}    & 0.8448 / \underline{\textbf{82.69s}}  & 0.8017 / \underline{\textbf{59.86s}}  & 0.7844 / 600.28s & 0.7931 / \underline{\textbf{119.64s}} \\
& \multicolumn{1}{|l|}{BiMax w/Mean} & \underline{\textbf{0.9353}} / \underline{\textbf{82.90s}}  & 0.8793 / \underline{\textbf{59.91s}}  & \underline{\textbf{0.9310}} / 598.58s & \underline{\textbf{0.9267}} / \underline{\textbf{134.87s}} \\
\hline
\hline
\multicolumn{6}{|l|}{\textbf{Experiments of Blob-o w/sent}~\small(F1 Score \(\uparrow\) / PPROC. Time~(sec.) \(\downarrow\))} \\
\hline
\multicolumn{1}{|c}{\multirow{4}{*}{ \makecell[c]{Blob \\ (Max~64) }}}
& \multicolumn{1}{|l|}{Mean-Pool}    & 0.8621 / 107.02s & \underline{\textbf{0.8663}} / 70.80s & \underline{\textbf{0.8750}} / \underline{\textbf{565.45s}} & \underline{\textbf{0.8448}} / 127.75s  \\
& \multicolumn{1}{|l|}{TK-PERT}      & 0.8663 / 164.87s & 0.8491 / 139.41s & 0.8534 / 655.54 & 0.8578 / 220.72s \\
& \multicolumn{1}{|l|}{OT w/Mean}    & \underline{\textbf{0.8233}} / 107.84s & \underline{\textbf{0.8405}} / 70.46s & 0.8362 / \underline{\textbf{564.84s}} & \underline{\textbf{0.8276}} / 128.12s \\
& \multicolumn{1}{|l|}{BiMax w/Mean} & 0.9009 / 106.65s & \underline{\textbf{0.9052}} / 71.16s & 0.9181 / \underline{\textbf{564.76s}} & \underline{\textbf{0.9052}} / 127.32s \\
\hline
\multicolumn{1}{|c}{\multirow{4}{*}{ \makecell[c]{Blob-o w/sent \\ (64, 4) }}}
& \multicolumn{1}{|l|}{Mean-Pool}    & 0.8707 / 108.21s & 0.8621 / 73.74s  & \underline{\textbf{0.8750}} / 600.28s & 0.8405 / 130.19s  \\
& \multicolumn{1}{|l|}{TK-PERT}      & \underline{\textbf{0.8879}} / 173.63s & \underline{\textbf{0.8578}} / 149.27s & \underline{\textbf{0.8621}} / 655.37s & \underline{\textbf{0.8664}} / 223.978s \\
& \multicolumn{1}{|l|}{OT w/Mean}    & \underline{\textbf{0.8233}} / 108.82s & 0.8362 / 73.87s  & \underline{\textbf{0.8405}} / 600.49s & \underline{\textbf{0.8276}} / 129.78s \\
& \multicolumn{1}{|l|}{BiMax w/Mean} & 0.9009 / 108.56s & 0.9009 / 73.63s  & 0.9138 / 600.78s & \underline{\textbf{0.9052}} / 130.08s \\
\hline
\multicolumn{1}{|c}{\multirow{4}{*}{ \makecell[c]{Blob-o w/sent \\ (128, 3) }}}
& \multicolumn{1}{|l|}{Mean-Pool}    & \underline{\textbf{0.8966}} / \underline{\textbf{84.83s}}  & 0.8189 / \underline{\textbf{59.92s}}  & 0.8621 / 578.21s & 0.8319 / \underline{\textbf{116.21s}}  \\
& \multicolumn{1}{|l|}{TK-PERT}      & \underline{\textbf{0.8879}} / \underline{\textbf{124.49s}} & 0.8362 / \underline{\textbf{108.65s}} & 0.8534 / \underline{\textbf{638.59s}} & 0.8578 / \underline{\textbf{179.93s}} \\
& \multicolumn{1}{|l|}{OT w/Mean}    & 0.7974 / \underline{\textbf{85.09s}}  & 0.7931 / \underline{\textbf{60.84s}}  & 0.7931 / 577.49s & 0.8060 / \underline{\textbf{116.06s}} \\
& \multicolumn{1}{|l|}{BiMax w/Mean} & \underline{\textbf{0.9095}} / \underline{\textbf{84.78s}}  & 0.8707 / \underline{\textbf{60.23s}}  & \underline{\textbf{0.9224}} / 577.78s & 0.8966 / \underline{\textbf{116.61s}} \\
\hline
\hline
\multicolumn{6}{|l|}{\textbf{Experiments of Blob-o w/tok-lim}~\small(F1 Score \(\uparrow\) / PPROC. Time~(sec.) \(\downarrow\))} \\
\hline
\multicolumn{1}{|c}{\multirow{4}{*}{ \makecell[c]{Blob \\ (Max~64) }}}
& \multicolumn{1}{|l|}{Mean-Pool}    & 0.8621 / 107.02s & \underline{\textbf{0.8663}} / 70.80s & \underline{\textbf{0.8750}} / \underline{\textbf{565.45s}} & \underline{\textbf{0.8448}} / 127.75s  \\
& \multicolumn{1}{|l|}{TK-PERT}      & 0.8663 / 164.87s & 0.8491 / 139.41s & 0.8534 / \underline{\textbf{655.54}} & 0.8578 / 220.72s \\
& \multicolumn{1}{|l|}{OT w/Mean}    & \underline{\textbf{0.8233}} / 107.84s & \underline{\textbf{0.8405}} / 70.46s & \underline{\textbf{0.8362}} / \underline{\textbf{564.84s}} & 0.8276 / 128.12s \\
& \multicolumn{1}{|l|}{BiMax w/Mean} & 0.9009 / 106.65s & \underline{\textbf{0.9052}} / 71.16s & 0.9181 / \underline{\textbf{564.76s}} & \underline{\textbf{0.9052}} / 127.32s \\
\hline
\multicolumn{1}{|c}{\multirow{4}{*}{ \makecell[c]{Blob-o w/tok-lim \\ (64, 0.15) }}}
& \multicolumn{1}{|l|}{Mean-Pool}    & 0.8621 / 111.63s & \underline{\textbf{0.8663}} / 74.65s  & \underline{\textbf{0.8750}} / 613.27s & 0.8405 / 130.53s  \\
& \multicolumn{1}{|l|}{TK-PERT}      & \underline{\textbf{0.8793}} / 175.39s & \underline{\textbf{0.8578}} / 148.23s & \underline{\textbf{0.8578}} / 672.91s & \underline{\textbf{0.8621}} / 225.43s \\
& \multicolumn{1}{|l|}{OT w/Mean}    & \underline{\textbf{0.8233}} / 111.41s & \underline{\textbf{0.8405}} / 74.97s  & \underline{\textbf{0.8362}} / 613.65s & \underline{\textbf{0.8319}} / 129.22s \\
& \multicolumn{1}{|l|}{BiMax w/Mean} & 0.9009 / 111.46s & \underline{\textbf{0.9052}} / 75.51s  & \underline{\textbf{0.9267}} / 613.19s & 0.9009 / 129.78s \\
\hline
\multicolumn{1}{|c}{\multirow{4}{*}{ \makecell[c]{Blob-o w/tok-lim \\ (128, 0.45) }}}
& \multicolumn{1}{|l|}{Mean-Pool}    & \underline{\textbf{0.8966}} / \underline{\textbf{81.94s}}  & 0.8319 / \underline{\textbf{58.93s}}  & 0.8578 / 593.52s & 0.8103 / \underline{\textbf{117.29s}}  \\
& \multicolumn{1}{|l|}{TK-PERT}      & 0.8707 / \underline{\textbf{122.81s}} & 0.8491 / \underline{\textbf{107.59s}} & 0.8491 / \underline{\textbf{654.17s}} & 0.8448 / \underline{\textbf{179.66s}} \\
& \multicolumn{1}{|l|}{OT w/Mean}    & 0.7974 / \underline{\textbf{82.56s}}  & 0.7672 / \underline{\textbf{58.97s}}  & 0.7931 / 593.12s & 0.8017 / \underline{\textbf{117.22s}} \\
& \multicolumn{1}{|l|}{BiMax w/Mean} & \underline{\textbf{0.9138}} / \underline{\textbf{82.15s}}  & 0.8836 / \underline{\textbf{58.80s}}  & \underline{\textbf{0.9267}} / 593.23s & 0.8750 / \underline{\textbf{116.90s}} \\
\hline
\end{tabular}
\end{adjustbox}
\caption{
The comparative results between the three Blob-o approaches and the original Blob method.
Each model's highest F1 scores and shortest preprocessing time are tagged in \textbf{bold} with \underline{underline}.}
\label{tab:blob-o}
\end{table*}

Based on the ablation analysis conducted by \citet{wang-etal-2024-document},
the overlapping rate has a notable impact on the accuracy of OFLS.
Therefore, we hypothesize that appropriately introducing overlapping parts between Blobs might contribute to improvement. We design the following three approaches:
\vspace{-0.3cm}

\hspace*{\fill}

\noindent
\textbf{Blob-o w/tok:} 
For any two given Blobs $A$ and $B$ and a specified ratio $r$,
we copy the last $len(A)\times r$ tokens from Blob $A$ to the beginning of Blob $B$ 
while simultaneously replicating the first $len(B)\times r$ tokens from Blob $B$ to the end of Blob $A$,
with all operations performed at the token level.
\vspace{-0.3cm}

\hspace*{\fill}

\noindent
\textbf{Blob-o w/sent:} 
For any two given Blobs $A$ and $B$ and a specified number $n$,
we copy the last $n$ sentences from Blob $A$ to the beginning of Blob $B$ 
while simultaneously replicating the first $n$ sentences from Blob $B$ to the end of Blob $A$,
with all operations performed at the sentence level.
\vspace{-0.3cm}

\hspace*{\fill}

\noindent
\textbf{Blob-o w/tok-lim:} 
For any two given Blobs $A$ and $B$ and a specified ratio $r$,
we copy the multiple sentences from the end of Blob $A$,
comprising no more than $len(A)\times r$ tokens,
to the beginning of Blob $B$,
while simultaneously replicating multiple sentences from the beginning of Blob $B$,
containing no more than $len(B)\times r$ tokens,
to the end of Blob $A$,
with all operations performed at the sentence level.

\begin{figure}[ht]
    \centering
    \includegraphics[scale=0.19]{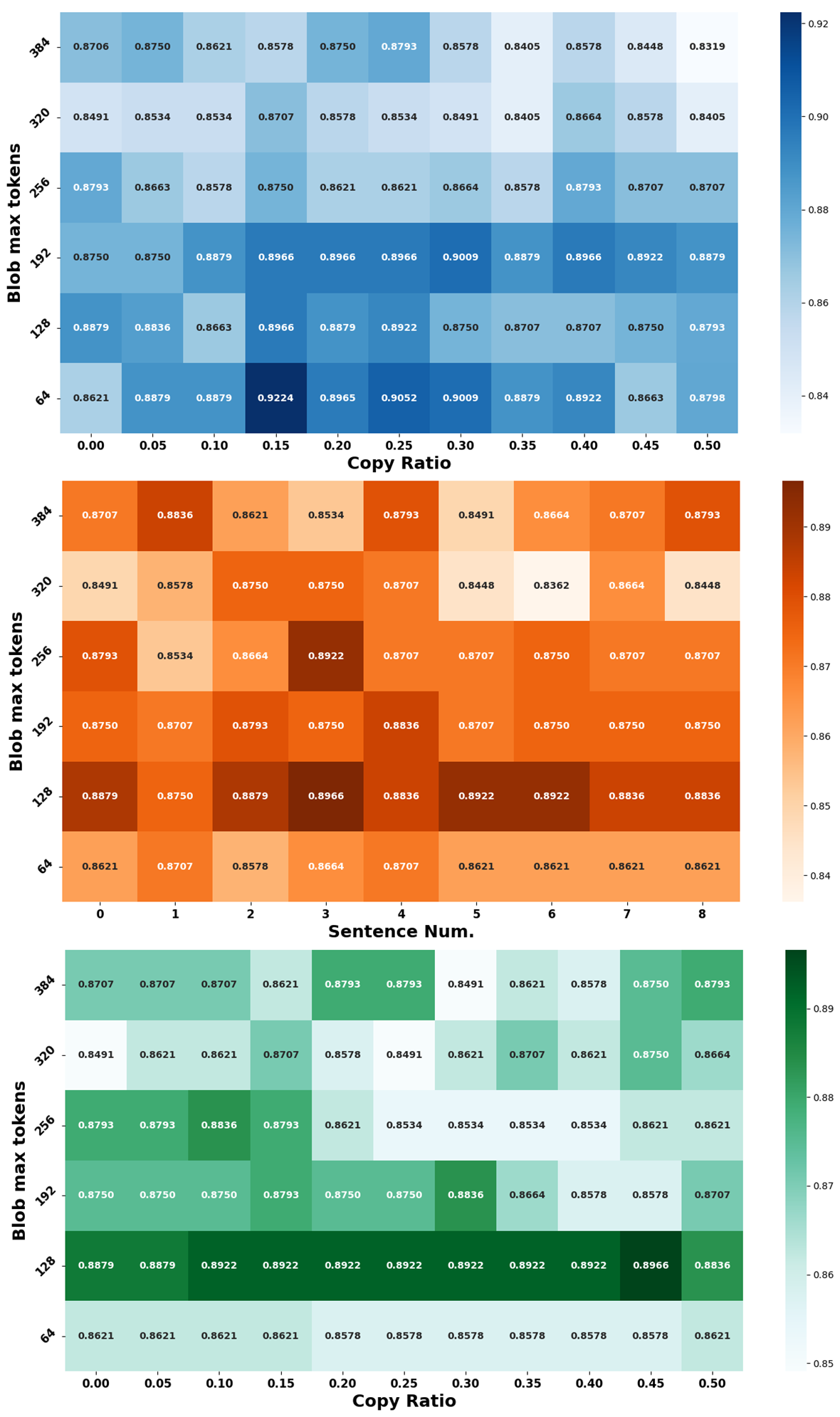}
    \caption{
    The preliminary experimental results of the three Blob-o approaches using the Mean-Pool method based on the LaBSE model. 
    From top to bottom, the results correspond to \textcolor{deepdeepblue}{Blob-o w/tok}, \textcolor{orange}{Blob-o w/sent}, and \textcolor{deepgreen}{Blob-o w/tok-lim}, respectively.}
    \label{fig:blob-o}
\end{figure}

We perform preliminary experiments using the Mean-Pool method based on the LaBSE model 
to examine appropriate combinations of the maximum token limitation for Blob composition 
and the hyperparameters associated with the aforementioned three approaches. 
The results of these experiments are presented in Figure~\ref{fig:blob-o}.

The exploration is limited to the LaBSE model and Mean-Pool method,
possibly creating bias if their optimal configurations are 
directly applied to alternative models and document alignment methods.
To address this, we implement a comparative evaluation by examining two cases from each Blob-o approach 
against the original Blob method.

As shown in Table~\ref{tab:blob-o}, 
except for the Blob-o w/tok~(64, 0.15) with the LaBSE model, 
which performs well compared to the original Blob~(Max 64), 
other cases do not exhibit significant improvement.
Combined with the results in Table~\ref{tab:blob},
it can be observed that the OT method does not integrate well with Blob, 
potentially because OT tends to achieve superior performance with finer segmentation granularity,
which contradicts the Blob approach.

The Blob method is initially introduced to our research 
to reduce the impact of boilerplate text,
preserve the meaning of source sentences, 
and accelerate embedding processes. 
However, on the MnRN Dataset, it shows little improvement compared to SBS. 
This could be attributed to multiple factors, 
such as the embedding model's potential inability to effectively represent long segments, 
the relatively small scale of the MnRN dataset, 
and the lack of coherence between Blobs. 
Nevertheless, the results indicate that even with the use of overlapping,
the performance of the Blob method on the MnRN dataset has not been enhanced overall.

\section{Detailed experimental results on the Fernando Dataset}
\label{sec:detail_low_rslt}
The detailed experimental results are reported in Table~\ref{tab:low_lang_detail_rslt}. 
Following \citet{ref1}, we record the precision, recall, 
and F1 score of each language pair under each web domain. 
Since only the Army and Hiru domains in the GitHub data provided by the authors match the data size reported in the original paper, 
we reproduce their strong baseline method ``GMD-SL''~(i.e., the GMD method with segment length as the weighting scheme) using the authors’ released code\footnote{\url{https://github.com/nlpcuom/parallel_corpus_mining/blob/master/document_alignment/GreedyMoversDistance.py}} for comparison. 
The ITN and NewsFirst domains are excluded from comparison due to substantial discrepancies in the data. 
Moreover, since OT in this paper adopts segment frequency as its weighting strategy~(OT-SF),
we include the results of ``OT-SL + SBS'' to ensure a fair comparison with GMD.

\begin{table*}[htbp]
\centering
\small
\begin{adjustbox}{width=0.93\textwidth}
\begin{tabular}{lcccccccccccc}
\hline
\multirow{3}{*}{LaBSE} & \multirow{3}{*}{\makecell[c]{Segment \\ Strategy }} & \multicolumn{11}{c}{\textbf{Army}} \\
\cline{3-13}
& & \multicolumn{3}{c}{En-Si} && \multicolumn{3}{c}{En-Ta} && \multicolumn{3}{c}{Si-Ta} \\
\cline{3-5} \cline{7-9}\cline{11-13}
& & R & P & F1 && R & P & F1 && R & P & F1 \\
\hline
\multirow{2}{*}{Mean-Pool}     & SBS  & 98.16 & 90.47 & 94.16 && 80.08 & 81.52 & 80.79 && 81.88 & 75.03 & 78.30 \\
                               & OFLS & 98.43 & 90.27 & 94.18 && 93.83 & 85.17 & 89.29 && 94.04 & 79.58 & 86.20 \\
\cdashline{2-13}
\multirow{2}{*}{TK-PERT}       & SBS  & 98.32 & 90.53 & 94.27 && 84.20 & 81.51 & 82.83 && 86.63 & 76.07 & 81.01 \\
                               & OFLS & 98.81 & 90.85 & 94.66 && 94.36 & 85.97 & 89.97 && 95.18 & 81.54 & 87.84 \\
\cdashline{2-13}
\multirow{2}{*}{OT-SF}         & SBS  & 98.05 & 90.46 & 94.11 && 83.49 & 84.38 & 83.93 && 85.36 & 77.73 & 81.37 \\
                               & OFLS & 98.97 & 90.81 & 94.72 && 96.24 & 87.17 & 91.48 && 97.34 & 82.32 & 89.20 \\
\cdashline{2-13}
\multirow{2}{*}{BiMax}         & SBS  & 98.32 & 90.71 & 94.37 && 85.49 & 86.30 & 85.89 && 87.58 & 79.75 & 83.48 \\
                               & OFLS & \textbf{99.30} & \textbf{91.12} & \textbf{95.03} && \textbf{97.46} & \textbf{88.14} & \textbf{92.52} && \textbf{97.72} & \textbf{82.48} & \textbf{89.50} \\
\hline
\makecell[l]{GMD-SL \\ (orig)} & SBS  & 99.73 & 94.85 & 97.23 && 98.47 & 91.89 & 95.07 && 99.11 & 86.84 & 92.57 \\
\cdashline{2-13}
\multirow{2}{*}{\makecell[l]{GMD-SL \\ (our)}}& SBS  & 98.43 & 90.72 & 94.42 && 86.02 & 87.83 & 86.42 && 88.59 & 80.62 & 84.42 \\
                                              & OFLS & \textbf{99.30} & 91.02 & 94.98 && 96.89 & 87.67 & 92.05 && 97.47 & 82.33 & 89.26 \\
\cdashline{2-13}
OT-SL                          & SBS  & 98.27 & 90.53 & 94.24 && 84.78 & 85.69 & 85.23 && 87.52 & 79.73 & 83.44 \\
\hline
& &  &  &  &&  &  & &&  &  & \\
\hline
\multirow{3}{*}{LaBSE} & \multirow{3}{*}{\makecell[c]{Segment \\ Strategy }} & \multicolumn{11}{c}{\textbf{Hiru}} \\
\cline{3-13}
& & \multicolumn{3}{c}{En-Si} && \multicolumn{3}{c}{En-Ta} && \multicolumn{3}{c}{Si-Ta} \\
\cline{3-5} \cline{7-9}\cline{11-13}
& & R & P & F1 && R & P & F1 && R & P & F1 \\
\hline
\multirow{2}{*}{Mean-Pool}     & SBS  & 89.33 & 76.38 & 82.35 && 73.59 & 52.69 & 61.41 && 77.02 & 57.47 & 65.83 \\
                               & OFLS & 88.69 & 75.83 & 81.76 && 75.81 & 54.28 & 63.27 && 77.87 & 57.93 & 66.44 \\
\cdashline{2-13}
\multirow{2}{*}{TK-PERT}       & SBS  & 80.46 & 68.79 & 74.17 && 58.89 & 42.24 & 49.20 && 66.18 & 49.85 & 56.87 \\
                               & OFLS & 82.03 & 70.13 & 75.62 && 65.90 & 47.18 & 54.99 && 69.73 & 51.84 & 59.47 \\
\cdashline{2-13}
\multirow{2}{*}{OT-SF}         & SBS  & 86.75 & 74.17 & 79.97 && 70.60 & 50.55 & 58.92 && 77.62 & 57.81 & 66.27 \\
                               & OFLS & 86.75 & 74.17 & 79.97 && 71.37 & 51.10 & 59.56 && 78.27 & 58.21 & 66.77 \\
\cdashline{2-13}
\multirow{2}{*}{BiMax}         & SBS  & \textbf{93.49} & \textbf{79.93} & \textbf{86.18} && 79.74 & 57.13 & 66.57 && 82.42 & 60.93 & 70.06 \\
                               & OFLS & 93.13 & 79.62 & 85.85 && \textbf{82.48} & \textbf{59.06} & \textbf{68.83} && \textbf{83.27} & \textbf{61.38} & \textbf{70.67} \\
\hline
\makecell[l]{GMD-SL \\ (orig)} & SBS  & 95.42 & 82.44 & 88.45 && 87.09 & 62.71 & 72.92 && 87.46 & 65.19 & 74.66 \\
\cdashline{2-13}
\multirow{2}{*}{\makecell[l]{GMD-SL \\ (our)}}& SBS  & 91.98 & 78.64 & 84.79 && 78.12 & 55.94 & 65.19 && 80.42 & 59.39 & 68.32 \\
                                              & OFLS & 89.05 & 76.13 & 82.09 && 78.55 & 56.24 & 65.66 && 79.87 & 58.87 & 67.78 \\
\cdashline{2-13}
OT-SL                          & SBS  & 88.33 & 75.52 & 81.43 && 72.22 & 51.71 & 60.27 && 78.12 & 58.23 & 66.72 \\
\hline
& &  &  &  &&  &  & &&  &  & \\
\hline
\multirow{3}{*}{LaBSE} & \multirow{3}{*}{\makecell[c]{Segment \\ Strategy }} & \multicolumn{11}{c}{\textbf{ITN}} \\
\cline{3-13}
& & \multicolumn{3}{c}{En-Si} && \multicolumn{3}{c}{En-Ta} && \multicolumn{3}{c}{Si-Ta} \\
\cline{3-5} \cline{7-9}\cline{11-13}
& & R & P & F1 && R & P & F1 && R & P & F1 \\
\hline
\multirow{2}{*}{Mean-Pool}     & SBS  & 85.51 & 15.69 & 26.52 && 74.11 & 5.98 & 11.07 && 91.18 & 2.05 & 4.02 \\
                               & OFLS & 83.24 & 15.47 & 26.09 && 83.04 & 6.78 & 12.54 && 91.18 & 2.04 & 4.00 \\
\cdashline{2-13}
\multirow{2}{*}{TK-PERT}       & SBS  & 78.69 & 14.40 & 24.34 && 72.32 & 5.80 & 10.74 && 85.29 & 1.92 & 3.75 \\
                               & OFLS & 80.40 & 14.81 & 25.01 && 79.46 & 6.36 & 11.78 && 79.41 & 1.78 & 3.48 \\
\cdashline{2-13}
\multirow{2}{*}{OT-SF}         & SBS  & 82.67 & 15.15 & 25.60 && 76.79 & 6.23 & 11.53 && 91.18 & 2.06 & 4.02 \\
                               & OFLS & 83.52 & 15.50 & 26.14 && 83.04 & 6.82 & 12.60 && 88.24 & 1.98 & 3.87 \\
\cdashline{2-13}
\multirow{2}{*}{BiMax}         & SBS  & \textbf{89.49} & \textbf{16.35} & \textbf{27.64} && 83.04 & 6.71 & 12.42 && \textbf{97.06} & \textbf{2.19} & \textbf{4.28} \\
                               & OFLS & 86.36 & 15.97 & 26.96 && \textbf{91.96} & \textbf{7.44} & \textbf{13.76} && 91.18 & 2.04 & 4.00 \\
\hline
\multirow{2}{*}{\makecell[l]{GMD-SL \\ (our)}}& SBS  & 88.07 & 16.09 & 27.20 && 85.71 & 6.99 & 12.92 && 94.12 & 2.12 & 4.14 \\
                                              & OFLS & 85.80 & 15.88 & 26.80 && 83.93 & 6.83 & 12.63 && 88.24 & 1.98 & 3.87 \\
\cdashline{2-13}
OT-SL                          & SBS  & 83.24 & 15.25 & 25.78 && 75.89 & 6.15 & 11.39 && 94.12 & 2.12 & 4.14 \\
\hline
& &  &  &  &&  &  & &&  &  & \\
\hline
\multirow{3}{*}{LaBSE} & \multirow{3}{*}{\makecell[c]{Segment \\ Strategy }} & \multicolumn{11}{c}{\textbf{NewsFirst}} \\
\cline{3-13}
& & \multicolumn{3}{c}{En-Si} && \multicolumn{3}{c}{En-Ta} && \multicolumn{3}{c}{Si-Ta} \\
\cline{3-5} \cline{7-9}\cline{11-13}
& & R & P & F1 && R & P & F1 && R & P & F1 \\
\hline
\multirow{2}{*}{Mean-Pool}     & SBS  & 88.95 & 18.29 & 30.34 && 81.33 & 12.93 & 22.31 && 84.54 & 4.67 & 8.85 \\
                               & OFLS & 90.99 & 18.17 & 30.28 && 86.71 & 13.54 & 23.43 && 87.63 & 4.82 & 9.14 \\
\cdashline{2-13}
\multirow{2}{*}{TK-PERT}       & SBS  & 87.79 & 17.71 & 29.48 && 81.01 & 12.84 & 22.17 && 82.47 & 4.53 & 8.58 \\
                               & OFLS & 86.92 & 17.27 & 28.82 && 82.59 & 12.84 & 22.22 && 82.47 & 4.53 & 8.60 \\
\cdashline{2-13}
\multirow{2}{*}{OT-SF}         & SBS  & 88.37 & 18.16 & 30.13 && 86.08 & 13.57 & 23.45 && 85.57 & 4.75 & 9.00 \\
                               & OFLS & 90.41 & 18.06 & 30.11 && 87.34 & 13.50 & 23.39 && 88.66 & 4.88 & 9.25 \\
\cdashline{2-13}
\multirow{2}{*}{BiMax}         & SBS  & \textbf{95.06} & 19.34 & \textbf{32.14} && 90.51 & \textbf{14.25} & \textbf{24.62} && 88.66 & 4.88 & 9.25 \\
                               & OFLS & 93.02 & \textbf{19.84} & 30.74 && \textbf{91.46} & 14.13 & 24.47 && \textbf{91.75} & \textbf{4.98} & \textbf{9.44} \\
\hline
\multirow{2}{*}{\makecell[l]{GMD-SL \\ (our)}}& SBS  & 93.02 & 18.95 & 31.48 && 90.19 & 14.20 & 25.54 && 88.66 & 4.87 & 9.23 \\
                                              & OFLS & 93.90 & 18.55 & 30.98 && \textbf{91.46} & 14.03 & 24.33 && 89.69 & 4.88 & 9.26 \\
\cdashline{2-13}
OT-SL                          & SBS  & 89.53 & 18.38 & 30.50 && 86.71 & 13.67 & 23.62 && 86.60 & 4.78 & 9.06 \\
\hline
\end{tabular}
\end{adjustbox}
\caption{The detailed results on the Fernando dataset.
We highlight the best one in \textbf{bold} for each column.
``GMD-SL~(orig)'' represents the results in the original paper~\citep{ref1}.
}
\label{tab:low_lang_detail_rslt}
\end{table*}

As shown in Table~\ref{tab:low_lang_detail_rslt}, 
even on the Army and Hiru domains, 
the results of GMD-SL reported in the original paper differ substantially from those reproduced in our experiments. 
This discrepancy is the main reason we do not compare our results directly with the original paper.
Then, BiMax consistently achieves the highest accuracy in all cases.
However, it can be observed that TK-PERT and OT, which performed well on the WMT16 test data,
do not achieve satisfactory results in this experiment.
This may be attributed to the fact that the Fernando dataset primarily consists of relatively short documents, 
whereas both methods are better suited for long texts. 
Moreover, this also reveals a limitation of the current version of TK-PERT: 
the number of windows per document is fixed,
which is clearly suboptimal for handling datasets with diverse length distributions, 
along with the challenge of properly configuring hyperparameters in advance. 
These inabilities remain an issue that TK-PERT could potentially improve upon.

As a greedy-search variant of OT, 
GMD achieves better performance than OT on the Fernando dataset, 
ranking second only to BiMax and clearly demonstrating its superior accuracy. 
However, due to its exhaustive traversal of all segment pairs, 
the computational cost grows significantly with the number of segments, 
leading to a marked slowdown, 
particularly in the case that OFLS divides documents into shorter segments. 
When running GMD using the implementation provided by~\citet{ref1}, 
its throughput under the OFLS setting was more than 20 times lower than BiMax and even more than 5 times lower than OT,
indicating that further improvements are needed at both the algorithmic and implementation levels.

\section{Alignment Accuracy Analysis based on Document Length}
\label{sec:ana_doclen}

\begin{figure*}[ht]
    \centering
    \includegraphics[scale=0.28]{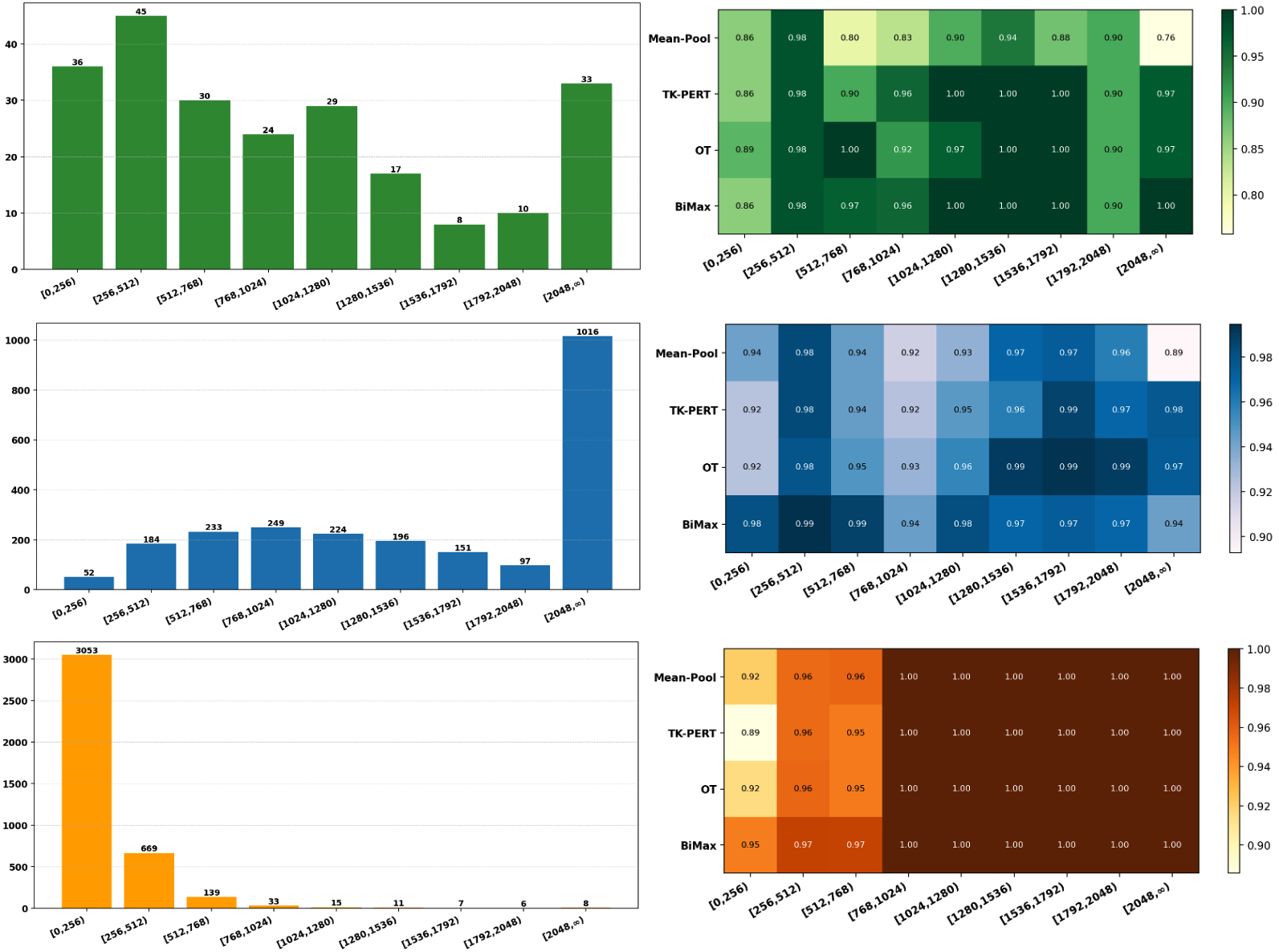}
    \caption{
    The length distribution of the English side of the gold data
    and the recall results of OFLS-based document alignment methods across different length intervals. 
    From top to bottom, the results correspond to the \textcolor{deepgreen}{MnRN dataset}, \textcolor{deepdeepblue}{WMT16 test data}, and \textcolor{orange}{Fernando dataset}, respectively.}
    \label{fig:len_distri}
\end{figure*}

Since English is the common language across the MnRN dataset, WMT16 test data, and the Fernando dataset\footnote{
For the Fernando dataset, 
we merge the En–Si data from the four web-domains for our analysis.
},
we examine the length distribution of the English documents in the gold data,
measuring text length by the number of tokens obtained after tokenization with the LaBSE model. 
Figure~\ref{fig:len_distri} presents the recall performance of the OFLS-based methods across different length intervals.

It can be observed that the length distributions of the gold data differ across the three datasets.
The MnRN dataset exhibits a relatively balanced distribution but contains more short documents; 
the WMT16 test data are primarily concentrated on long documents; 
in contrast, the Fernando dataset is almost entirely composed of short texts, 
with virtually no long documents.

Overall, embedding-based document alignment methods tend to perform less effectively on short texts,
particularly for documents in the $[0, 256)$ length interval.
Among them, BiMax demonstrates the strongest capability in handling short documents, 
achieving a recall of $0.95$ in the $[0, 256)$ interval of the Fernando dataset. 
Nevertheless, this performance is still lower than the accuracy achieved by
TK-PERT and OT on long documents in the $[2048, \infty)$ interval of the WMT16 test data.
A potential reason for this, as revealed through our examination of the MnRN dataset, 
is that even short documents often contain a substantial amount of boilerplate text, 
which appears repeatedly across documents within the same web domain.
This repetition further reduces the space for discriminative content, 
making it difficult for sentence-level embedding methods to capture fine-grained features.

As we noted in Section~\ref{sec:low_lang} and Appendix~\ref{sec:detail_low_rslt}, 
the strong performance of TK-PERT and OT on the WMT16 test data 
is largely attributable to their effectiveness in handling long documents,
an aspect where BiMax is comparatively weaker.
Conversely, these two methods perform less effectively than BiMax on the short-text Fernando dataset. 
This suggests that, rather than relying on a single alignment method, 
it may be worthwhile to consider a system that leverages different alignment approaches within the length intervals where each performs best.

\section{Downstream MT work for Document Alignment on the WMT23 Data Task}
\label{sec:mt_bench}
Up to date, there exist several datasets for evaluating document alignment tasks
~(e.g., WMT16 document alignment task~\citep{buck-koehn-2016-findings}, 
CC-Aligned Dataset~\citep{el-kishky-etal-2020-ccaligned}). 
However, the evaluations typically measure the accuracy of 
document alignment methods using recall or F1 scores on document pairs.
There has not yet been a publicly available system that evaluates document alignment accuracy 
through machine translation performance as a downstream task.

The WMT23 parallel data curation shared task~(WMT23 data task)~\citep{sloto-etal-2023-findings} focuses on identifying the best MT training data from provided web-crawled data,
including both document and sentence levels.
The final developed datasets are evaluated using a unified end-to-end MT system. 
As one of the participants, 
\citet{steingrimsson-2023-sentence} employed the document alignment method for one part of dataset creation,
ultimately combining it with the dataset via multi-filtering techniques to produce the final dataset. 
However, he did not explore their document alignment methodology in depth.

We hope to develop a comparative benchmark on the Estonian-Lithuanian~(et-lt) WMT23 data task and 
establish an end-to-end system for the document alignment task 
that utilizes machine translation~(MT) accuracy as the evaluation metric.
However, the process of converting document-aligned data into a parallel corpus involves multiple steps, 
such as sentence alignment and sentence pair filtering.
It can be anticipated that these steps will increase the permissible error margin for document alignment methods,
which means that when different document alignment methods achieve sufficiently high accuracy, 
the resulting datasets may not exhibit significant differences in quality.

\subsection{Procedure for Hierarchical Data Curation}
\label{sec:hier_datacur}
\subsubsection{Preprocessing with CH Data}
First, since documents from different hostnames~(web domains) are unlikely to be translations of each other,
we extract common hostnames that appear in both [\textit{documents.et.tsv}] and [\textit{documents.lt.tsv}] files provided by the WMT23 data task.
We then perform the following two preprocessing steps:
(1) Since documents may contain the same content even with different document IDs~(docids),
in order to conserve computational and storage resources during the subsequent embedding process and to reduce redundant sentence pairs in the final parallel dataset,
we deduplicate the source and target documents, respectively, within each hostname.
While cross-hostname duplicates also exist, 
we restrict deduplication to within-hostname operations to prevent certain hostnames from being completely depleted of documents.
(2) We remove exceptionally long documents\footnote{
For instance, for the hostname “lt.airbnb.com”, under LaBSE tokenization, 
since the longest document on the Estonian side does not exceed 1,000 tokens, we remove documents from the Lithuanian side that contain more than 10,000 tokens.
}, 
specifically, those whose length exceeds ten times the maximum length of documents in the opposed language, 
which can reasonably be assumed to lack aligned counterparts.
Following the steps described above, we divide the resulting Common Hostname Data~(CH Data) into two categories:
\begin{itemize}
    \item Common Hostname Data 1~(CH Data~1): Hostname data that have only one document on both the Estonian and Lithuanian sides.
    \item Common Hostname Data 2~(CH Data~2): Hostname data for which at least one side (et or lt) contains multiple documents.
\end{itemize}

We collect some information about the CH Data and record it in Table~\ref{tab:ch_data}.
Each document is stored in the format “URL\textbackslash t Hostname\textbackslash t 
docid\textbackslash t content~(encoded in base64)”.

\begin{table}[ht]
\centering
\begin{adjustbox}{width=0.47\textwidth}
\begin{tabular}{lccc}
\hline
                    & CH Data~1 & CH Data~2 & CH Data \\
\hline
Hostname num.       & 6,791     & 17,529    & 24,320  \\
Estonian Docs.      & 6,791     & 419,152   & 425,943 \\
Lithuanian Docs.    & 6,791     & 393,742   & 400,533 \\
\hline
\end{tabular}
\end{adjustbox}
\caption{Some statistical information of CH Data.}
\label{tab:ch_data}
\end{table}

\subsubsection{Document alignment}
Next, we perform document alignment between the et-lt documents.
We directly compute the similarity for CH Data~1 since each hostname can contain only a single document pair, 
and for CH Data~2, we perform retrieval. 
The final training and evaluation scripts provided by the WMT23 data task focus on the et-lt direction, 
so we follow them to set the retrieval direction as et-lt.
Subsequently, we merge the results from the two document alignment processes 
to obtain the CH document pairs~(CH docpairs) and the similarity score for each pair.

\subsubsection{Document-level Filtering}
\label{sec:doc_flt}
Because we perform deduplication only within each hostname for CH Data,
the resulting CH docpairs may still contain repeated content, 
and we cannot fully ensure that all documents are genuinely in Estonian or Lithuanian. 
Hence, we apply document-level filtering as follows:

\vspace{0.15cm}
\noindent \textbf{(I) Deduplication}: We sort the CH docpairs by similarity scores in descending order.
If an Estonian or Lithuanian document reappears in a later pair,
we remove that occurrence. 
In other words, we retain only the pairing with the highest similarity score 
for each document to eliminate duplicates.

\vspace{0.15cm}
\noindent \textbf{(II) Language identification}: Using the FastText model\footnote{
\url{https://fasttext.cc/docs/en/language-identification.html}} \cite{joulin2016fasttext, joulin2016bag},
we identify the language of each document in the remaining pairs from (I). 
We only preserve pairs whose source document is most likely Estonian (et)
and whose target document is most likely Lithuanian (lt).

\vspace{0.15cm}
Since the number of docpairs developed from different document alignment methods varies,
with the goal of comparing these methods, from a fairness standpoint,
a fair comparison would require setting a similarity threshold or fixing the sampling size to extract docpairs. However, because the similarity scales produced by methods~(e.g., OT and BiMax) differ substantially, 
adopting a single fixed threshold is impossible.
Consequently, we select a specified number of top-ranked docpairs (based on similarity) from the document-level filtering results for the subsequent sentence alignment.

\subsubsection{Sentence Alignment}
We use Vecalign \cite{thompson-koehn-2019-vecalign} to perform sentence alignment 
on the docpairs obtained in Section~\ref{sec:doc_flt}. 
Differing from the default settings, 
we set the overlap to 4 and replace the embedding from LASER to LaBSE.
Furthermore, using each sentence’s index in the document, 
we find the corresponding sentence ID~(sentid) in the file we compiled,
which is limited to the Common Hostname part from the
[\textit{sentences.et.tsv}] and [\textit{sentences.lt.tsv}] provided by the WMT23 data task.

\subsubsection{Sentence-level Filtering}
\label{sec:sent_flt}
In this step, we do not propose or employ any novel or complex methodology.
Instead, we carry out the necessary removal with the test and development data, 
as well as quality-based filtering of sentence pairs~(sentpairs):

\vspace{0.15cm}
\noindent \textbf{(III)Test\&Dev Removal}: Relying on the organizer-provided [\textit{exclude\_sent\_ids\_et-lt.txt}],
we remove all sentpairs whose sentid covers with any ID listed in this file.

\vspace{0.15cm}
\noindent \textbf{(IV) Quality-Based Filtering}: Following the approach of Steingrimsson \cite{steingrimsson-2023-sentence}, 
we retain only those pairs in which both the Estonian and Lithuanian sentences have more than three tokens 
~(tokenized simply by space).
We then use the LaBSE model for embedding and compute the cosine similarity for each sentpair, 
removing any pairs with a score below 0.4.

\vspace{0.15cm}
Similarly to Section~\ref{sec:doc_flt},
we sort the filtered sentpairs in descending order of cosine similarity 
and extract a fixed number of pairs as our final parallel dataset for training.

\subsection{Downstream MT Benchmark}

\begin{figure*}[ht]
    \centering
    \begin{subfigure}{0.488\textwidth}
        \centering
        \includegraphics[width=\linewidth]{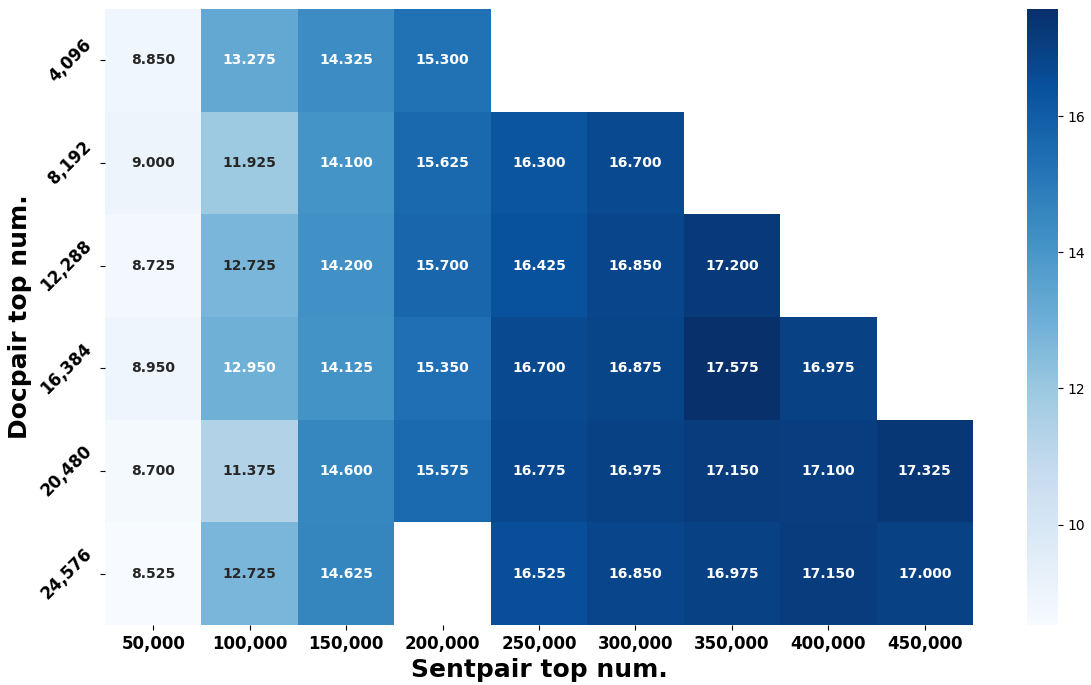} 
    \end{subfigure}
    \hfill
    \begin{subfigure}{0.488\textwidth}
        \centering
        \includegraphics[width=\linewidth]{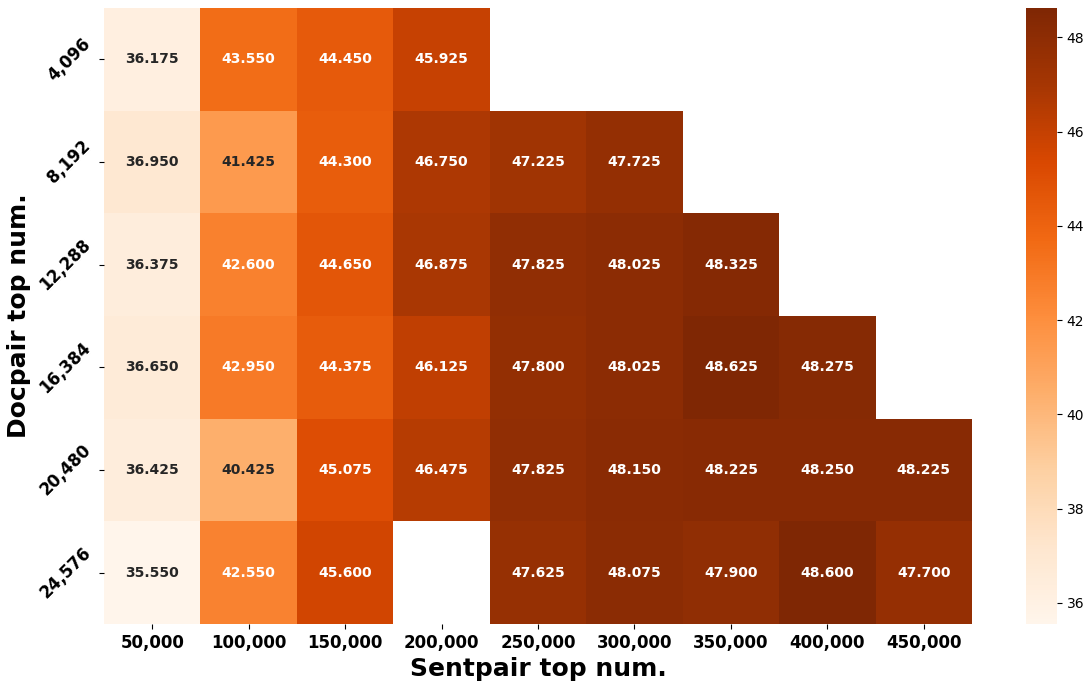} 
    \end{subfigure}
    \caption{Results of Avg BLEU~(left) and Avg ChrF~(right) for Mean-Pool with OFLS~(40, 0.5).}
    \label{fig:fig_mp}

    \vspace{1em}

    \begin{subfigure}{0.488\textwidth}
        \centering
        \includegraphics[width=\linewidth]{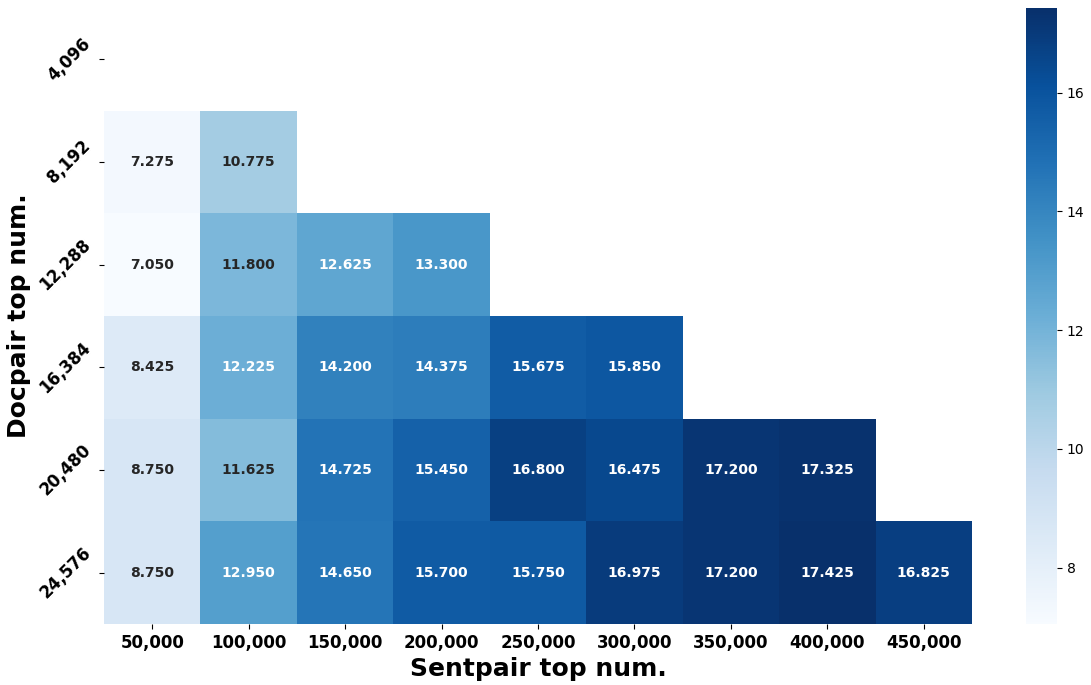}
    \end{subfigure}
    \hfill
    \begin{subfigure}{0.488\textwidth}
        \centering
        \includegraphics[width=\linewidth]{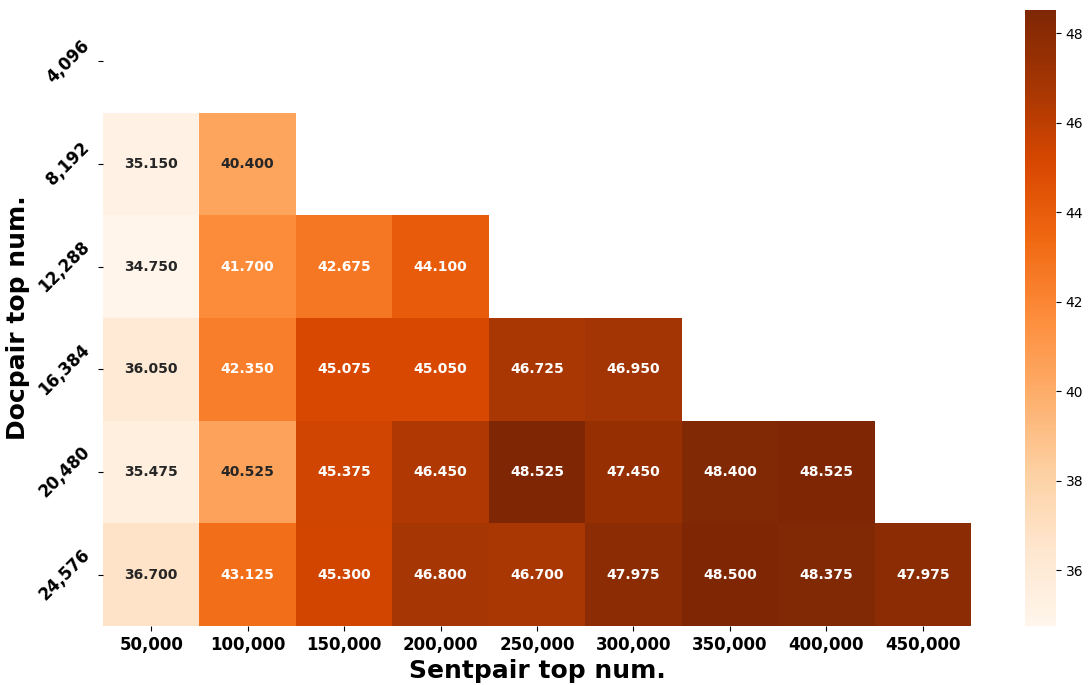} 
    \end{subfigure}
    \caption{Results of Avg BLEU~(left) and Avg ChrF~(right) for OT w/Mean with OFLS~(40, 0.5).}
    \label{fig:fig_ot}
\end{figure*}

\subsubsection{Experiment Settings}
As in our experiments on the WMT16 document alignment task and the MnRN dataset, 
we employ four document alignment methods,
Mean-Pool, TK-PERT, OT, and BiMax,
as well as two segmentation strategies, SBS and OFLS. 
The difference is that we rely solely on Mean-Pool to retrieve candidates for OT and BiMax, and we set the fixed-length to 40 and the overlapping rate to 0.5 for OFLS~(OFLS~(40, 0.5)).

We use the provided scripts to conduct end-to-end training, applying an early-stopping criterion that terminates training if the validation perplexity~(PPL) does not improve for 12 consecutive epochs. 
Then, the checkpoint with the lowest PPL is selected as the best model. 
However, the procedure for model selection will be discussed in more detail in Section~\ref{sec:dev_bench}.

The evaluation uses BLEU \cite{papineni-etal-2002-bleu} and chrF \cite{popovic-2015-chrf} across four domains\footnote{
The organizers add EUconst as an additional held-out domain in the Findings paper \cite{sloto-etal-2023-findings} as part of the held-out test set, which has not been publicly available.
}
: EMEA, EUbookshop~(EUB), Europarl~(EP), and JRC-Acquis~(JRC).
All datasets are released by OPUS\footnote{\url{https://opus.nlpl.eu/}} \cite{tiedemann-2012-parallel}.

We conduct all experiments except training on two H100 GPUs,
while the NMT model training is done on one A6000 GPU\footnote{
At the time this study was finished,
unfortunately, the Sockeye Python library used by the training script was not yet compatible 
with Torch version 2.0 or higher, 
which is necessary for H100 GPUs. 
Thus, we migrated model training to one A6000 GPU. 
However, because the original script assumes the use of eight V100 GPUs, 
we increased the batch size to eight times its original size.
}.

\subsubsection{Selection for Docpairs and Sentpairs}
\label{sec:select_num}
\begin{table*}[ht]
\centering
\resizebox{\textwidth}{!}{
\begin{tabular}{|l|c|c|c|c|c|c|}
\hline
\multirow{3}{*}{\textbf{\makecell[c]{ Document \\ Alignment }}} 
& \multirow{3}{*}{\textbf{\makecell[c]{ Segmentation \\ Strategy }}} 
& \multicolumn{2}{c|}{\textbf{Sim. Time~(sec.)}}     & \multicolumn{3}{c|}{\textbf{Pair num.}}     \\ 
\cline{3-7} 
    &      & \textbf{CH Data 1} & \textbf{CH Data 2} & \textbf{\makecell[c]{ Orig. \\ Docpairs }} & \textbf{\makecell[c]{ Doc-Flt \\ Docpairs }} & \textbf{\makecell[c]{ Vecalign \\ (top 24,586 docpairs) } + \makecell[c]{ Sent-Flt \\ Sentpairs }} \\ 
\hline
\multirow{2}{*}{Mean-Pool}
    & SBS             & 112.03s & 10148.23s & 78,539 & 33,627 $\Rightarrow$ 24,586 & 493,024 $\rightarrow$ 400,000 \\ 
    & OFLS~(40, 0.5)  & 94.55s  & 8043.31s  & 79,362 & 34,173 $\Rightarrow$ 24,586 & 490,693 $\rightarrow$ 400,000 \\ 
\hline
\multirow{2}{*}{TK-PERT}     
    & SBS             & 125.25s & 10646.20s & 78,785 & 33,722 $\Rightarrow$ 24,586 & 501,745 $\rightarrow$ 400,000 \\ 
    & OFLS~(40, 0.5)  & 107.43s & 8750.86s  & 79,405 & 34,138 $\Rightarrow$ 24,586 & 496,471 $\rightarrow$ 400,000 \\ 
\hline
\multirow{2}{*}{OT w/Mean}     
    & SBS             & 177.32s & 25761.43s & 78,655 & 33,306 $\Rightarrow$ 24,586 & 491,069 $\rightarrow$ 400,000 \\  
    & OFLS~(40, 0.5)  & 144.86s & 20416.14s & 79,466 & 33,802 $\Rightarrow$ 24,586 & 483,219 $\rightarrow$ 400,000 \\ 
\hline
\multirow{2}{*}{BiMax w/Mean}     
    & SBS             & 143.13s & 13018.73s & 78,694 & 33,818 $\Rightarrow$ 24,586 & 456,492$\rightarrow$ 400,000 \\
    & OFLS~(40, 0.5)  & 120.34s & 10472.15s & 79,463 & 34,228 $\Rightarrow$ 24,586 & 455,765 $\rightarrow$ 400,000 \\ 
\hline
\end{tabular}
}
\caption{Some information on the data development process,
“Sim. Time” represents the time cost by the similarity calculation,
“Orig. Docpairs” represents the docpairs before document-level filtering,
“Doc-Flt Docpairs” represents the docpairs after document-level filtering,
“Vecalign~(top 24,586 docpairs) + Sent-Flt Sentpairs” represents 
the sentpairs after sentence-level filtering using top 24,586 docpairs.}
\label{tab:info_benchmark}
\end{table*}

Before conducting experiments on all combinations of document alignment methods and segmentation strategies,
we should determine the number of docpairs and sentpairs required for our benchmark,
as described in Sections~\ref{sec:doc_flt} and \ref{sec:sent_flt}. 
Therefore, we select Mean-Pool and OT with the OFLS~(40, 0.5) as representatives,
gradually reducing the number of docpairs from 24,576 in increments of 4,096,
and running experiments in the step of 50,000 sentpairs for each docpair number.
Note that after document-level filtering, although Mean-Pool and OT with OFLS~(40, 0.5) retain 34,173 and 33,802 pairs, respectively,
at the 24,576 point, the docpair similarities for Mean-Pool and OT reach 0.75 and 0.32~(in the range [0,1]),
with OT’s similarity value already considered very low.

Since presenting the BLEU and ChrF scores separately for all four domains would yield an enormous volume of data and complicate visualization, 
we use the average BLEU~(avg BLEU) and average ChrF~(avg ChrF) across the four domains, and the results are shown in Figure~\ref{fig:fig_mp} and \ref{fig:fig_ot}.

As shown in Figure~\ref{fig:fig_mp}, 
at the point of docpair top num. 24,576 and sentpair top num. 200,000, 
Mean-Pool with OFLS~(40, 0.5) yields no data because training failed~(possibly due to overfitting). Comparing the two document alignment methods reveals
that Mean-Pool provides more sentpairs than OT. 
Meanwhile, for both methods, regardless of the docpair count, 
avg BLEU and avg ChrF display a clear upward trend until the sentpair top number reaches 350,000, 
after which they level off.
Therefore, we decide to use \textbf{docpair top num. 24,576} and \textbf{sentpair top num. 400,000} as our benchmark settings for two reasons as follows:
\begin{itemize}
    \item Mean-Pool and OT methods roughly achieve their highest accuracy at around 400,000 sentpairs.
    \item Considering methods like OT that probably generate fewer sentence pairs, 
    we avoid choosing the maximum possible number of sentpairs for each docpair top number~(e.g., docpair top num. 24,576 and sentpair top num. 450,000).
\end{itemize}

\subsubsection{Developing MT Benchmark for various method combinations}
\label{sec:dev_bench}
In this section, we adopt docpair top num. 24,576 and sentpair top num. 400,000, 
which are determined in Section~\ref{sec:select_num},
and follow the hierarchical data curation procedures described in Section~\ref{sec:hier_datacur}.
Under these settings, we conduct experiments on all combinations of the four document alignment methods
and the two segmentation strategies, 
and record some details of the data development process in Table~\ref{tab:info_benchmark}.

It is apparent that as the number of docpairs increases, 
compared to the small-scale MnRN dataset,
the similarity calculation time gap grows significantly.
Consequently, under both SBS and OFLS conditions for CH Data 2, 
OT takes nearly twice as long as BiMax, 
whereas Mean-Pool is still the fastest method.

Next, we performed five replicate experiments for each of the methods that produced 400,000 sentpairs in Table~\ref{tab:info_benchmark},
and the results are presented in Table~\ref{tab:benchmark_result}.
However, we do not rely on the original approach of selecting the best checkpoint solely based on the validation perplexity~(PPL).
We observe that when the dataset size is relatively small, 
the valid PPL converges more quickly than other metrics, such as BLEU, ChrF, and Rouge-L, indicating that the checkpoint selected exclusively by PPL is unreasonable.
Therefore, instead of relying on PPL alone, 
we sum the rankings for PPL, BLEU, ChrF, and Rouge-L on the validation data to determine the best checkpoint. 
In cases where multiple checkpoints yield the same total ranking, 
actually any of those checkpoints can be chosen.
Nonetheless, we impose a priority order of BLEU > ChrF > PPL > Rouge-L to select the final best model,
and this selection approach is determined as Auto-Rank.

As shown in Table~\ref{tab:benchmark_result}, 
considering both BLEU and ChrF,
there are no substantial differences among the document alignment methods or between the SBS and OFLS segmentation strategies, 
except for a slight advantage exhibited by BiMax over the other three methods in the EP and JRC-Acquis domains. 
This phenomenon may be attributed to multiple factors: 
the process from document alignment to final dataset construction involves numerous intermediate stages, 
and variability introduced at any of these stages may contribute to the observed homogenization of results.

\begin{table*}[ht]
\centering
\resizebox{\textwidth}{!}{
\begin{tabular}{|l|c|c|c|c|c|c|c|c|c|}
\hline
\multirow{2}{*}{\textbf{\makecell[c]{ Document \\ Alignment }}} 
& \multirow{2}{*}{\textbf{\makecell[c]{ Segmentation \\ Strategy }}} 
& \multicolumn{4}{c|}{\textbf{BLEU}}     
& \multicolumn{4}{c|}{\textbf{ChrF}}     \\ 
\cline{3-10} 
                                       &                                    & \textbf{EMEA} & \textbf{EUB} & \textbf{EP}  & \textbf{JRC} & \textbf{EMEA} & \textbf{EUB} & \textbf{EP}  & \textbf{JRC} \\ 
\hline
\multirow{2}{*}{Mean-Pool}
    & SBS          & 14.7$\pm$0.1 & 17.7$\pm$0.1 & 15.7$\pm$0.2 & 23.9$\pm$0.1 & 44.3$\pm$0.1 & 50.4$\pm$0.1  & 49.0$\pm$0.1 & 53.6$\pm$0.2                 \\ 
    & OFLS~(40, 0.5)  & 14.7$\pm$0.2 & 17.8$\pm$0.2 & 15.7$\pm$0.2 & 23.9$\pm$0.1 & 44.4$\pm$0.3 & 50.6$\pm$0.2 & 49.0$\pm$0.2 & 53.7$\pm$0.2         \\ 
\hline
\multirow{2}{*}{TK-PERT}     
    & SBS          & 14.6$\pm$0.1 & 17.7$\pm$0.3 & 15.6$\pm$0.2 & 23.8$\pm$0.2 & 44.3$\pm$0.2 & 50.4$\pm$0.3  & 49.0$\pm$0.1 & 53.6$\pm$0.2         \\ 
    & OFLS~(40, 0.5)  & 14.6$\pm$0.1 & 17.7$\pm$0.2 & 15.7$\pm$0.1 & 23.9$\pm$0.2 & 44.3$\pm$0.3 & 50.4$\pm$0.3  & 49.0$\pm$0.2 & 53.6$\pm$0.3         \\ 
\hline
\multirow{2}{*}{OT w/Mean}     
    & SBS          & 14.6$\pm$0.1 & 17.8$\pm$0.3 & 15.6$\pm$0.2 & 23.9$\pm$0.1 & 44.5$\pm$0.1 & 50.5$\pm$0.3 & 49.0$\pm$0.2 & 53.6$\pm$0.2         \\  
    & OFLS~(40, 0.5)  & 14.6$\pm$0.1 & 17.8$\pm$0.2 & 15.6$\pm$0.2 & 23.9$\pm$0.1 & 44.5$\pm$0.1 & 50.5$\pm$0.3 & 49.0$\pm$0.2 & 53.6$\pm$0.2         \\ 
\hline
\multirow{2}{*}{BiMax w/Mean}     
    & SBS          & 14.7$\pm$0.1 & 17.6$\pm$0.1 & 15.9$\pm$0.1 & 24.0$\pm$0.2 & 44.4$\pm$0.1 & 50.5$\pm$0.1  & 49.3$\pm$0.1 & 53.8$\pm$0.1                 \\
    & OFLS~(40, 0.5)  & 14.7$\pm$0.1 & 17.8$\pm$0.1 & 15.9$\pm$0.1 & 24.0$\pm$0.3 & 44.5$\pm$0.2 & 50.5$\pm$0.1  & 49.3$\pm$0.3 & 53.9$\pm$0.4                 \\ 
\hline
\end{tabular}
}
\caption{Auto-Rank: Docpair top 24,586, Sentpair top 400,000 Performance.}
\label{tab:benchmark_result}
\end{table*}

\begin{table*}[ht]
\centering
\resizebox{\textwidth}{!}{
\begin{tabular}{|l|r|r|r|r|r|r|r|r|r|}
\hline
\multirow{2}{*}{\textbf{\makecell[c]{ Document \\ Alignment }}} & \multirow{2}{*}{\textbf{Sentpair Num.}} & \multicolumn{4}{c|}{\textbf{Bleu}}     & \multicolumn{4}{c|}{\textbf{ChrF}}     \\ 
\cline{3-10} 
                                       &                                    & \textbf{EMEA} & \textbf{EUB} & \textbf{EP}  & \textbf{JRC} & \textbf{EMEA} & \textbf{EUB} & \textbf{EP}  & \textbf{JRC} \\ 
\hline
Baseline \cite{sloto-etal-2023-findings}                          & 2,654,090 & 18.1 & 20.1 & 18.4 & 25.7 & 49.4 & 53.0 & 52.1 & 55.7         \\
\hline
Baseline~(Github)                           & 2,654,090 & 18.3 & 19.1 & 18.1 & 24.3 & 49.7 & 52.3 & 51.8 & 55.2         \\ 
Baseline \cite{minh-cong-etal-2023-fast}    & 2,654,090 & 18.3 & 19.1 & 18.1 & 24.3 & 49.7 & 52.3 & 51.8 & 55.2         \\ 
Baseline \cite{steingrimsson-2023-sentence} & 2,654,090 & 18.2 & 19.1 & 17.8 & 24.3 & 49.5 & 52.2 & 51.5 & 54.8         \\ 
Baseline~(Our)                               & 2,654,090 & 18.3 & 19.1 & 18.1 & 24.4 & 49.6 & 52.3 & 51.8 & 55.2         \\ 
\hline
\hline
\citet{minh-cong-etal-2023-fast}      & 12,918,719& 18.5 & 20.4 & 19.1 & 25.8 & 48.9 & 52.5 & 52.5 & 55.5         \\
\citet{steingrimsson-2023-sentence}   & 3,902,740 & 20.4 & 20.2 & 18.7 & 25.4 & 51.4 & 52.8 & 52.0 & 54.9         \\
Margin Score 3.2M \cite{sloto-etal-2023-findings}           & 3.2M      & 21.5 & 22.4 & 20.2 & 27.9 & 52.5 & 54.7 & 53.4 & 57.8         \\
\hline
\hline
Mean-Pool~(OFLS)$_{|addbase}$   & 2,992,080    & 19.2 & 20.3 & 18.1 & 26.6 & 50.6 & 53.4 & 51.3 & 56.5         \\ 
TK-PERT~(OFLS)$_{|addbase}$     & 2,997,590    & 19.2 & 20.4 & 18.0 & 26.4 & 50.5 & 53.1 & 51.1 & 56.1         \\ 
OT w/Mean~(OFLS)$_{|addbase}$   & 2,985,377    & 19.2 & 20.2 & 18.7 & 26.5 & 50.5 & 53.2 & 52.1 & 56.4         \\
BiMax w/Mean~(OFLS)$_{|addbase}$& 2,958,214    & 19.2 & 20.4 & 18.3 & 26.7 & 50.4 & 53.3 & 51.7 & 56.4         \\
\hline
\end{tabular}
}
\caption{The results of “document alignment methods + baseline” compared to previous works.}
\label{tab:com_pre}
\end{table*}

\subsubsection{Developing MT Benchmark compared to Previous Work}
\label{sec:dev_bench2}
In addition to comparing the various methods among themselves, 
we also aim to compare our results against those of the WMT23 data task participants and organizers.
However, the dataset we construct via document alignment and hierarchical mining can only serve as a high-quality but small-scale dataset,
and we still lack a large-scale base dataset.
Since we do not explore sentence filtering methods in depth,
we utilize the organizers’ baseline dataset~\citep{sloto-etal-2023-findings},
which consists of sentence pairs obtained by taking the top-1 cosine similarity from the LASER embeddings,
as our base dataset.
We then augment it with the dataset we create. 
We prioritize our dataset by removing duplicates from the baseline dataset and including all sentence pairs derived from the 24,586 docpairs.
Moreover, we rely exclusively on perplexity~(PPL) to determine the best checkpoint.

As the results shown in Table~\ref{tab:com_pre},
first, in comparison with the baseline method, 
we add less than one-fifth of its data size yet achieve a substantial improvement in accuracy 
in the EMEA, EUBookshop, and JRC-Acquis domains,
indicating the high quality of our document-alignment-derived dataset. 
Second, compared with other participants,
we achieve comprehensive high BLEU and ChrF in the JRC-Acquis domain compared to two participants,
and also outperform \citet{minh-cong-etal-2023-fast} in the EMEA domain.
However, we observe that the organizers’ baseline results on the 
EUbookshop and JRC-Acquis domains are substantially higher than both ours and those of other participants,
likely due to differences in system environments or some other reasons.
Accordingly, we refrain from direct comparison with their reported numbers~\citep{sloto-etal-2023-findings}. 
Nonetheless, given that their pipeline employs margin scores
for translation data mining—providing a clear performance advantage—
we hypothesize that replacing our base dataset with one extracted using margin scores may further enhance our results.

\subsection{Summarization of MT Benchmark for Document Alignment on the WMT23 Data Task}

In Appendix~\ref{sec:mt_bench}, we aim to develop an end-to-end system for the WMT23 data task, 
evaluating document alignment quality through its impact on downstream machine translation (MT) performance. 
We endeavor to present the development process with transparency and rigor. 
However, the final results exhibit a high degree of homogenization across methods~(as shown in Table~\ref{tab:benchmark_result}).
As discussed in Appendix~\ref{sec:dev_bench},
numerous intermediate variables exist in the process from document alignment to the construction of the final parallel sentence pair dataset. 
Additionally, factors such as the limited size of the Common Hostname data and 
the characteristics of the Estonian-Lithuanian language pair likely 
contribute to deviated results from our expectations.
Furthermore, due to the absence of ground-truth document pairs in the WMT23 data task,
the expected comparison of the four alignment methods is based 
on their performance on the MnRN dataset and the WMT16 test data;
consequently, we cannot draw definitive conclusions about 
their relative effectiveness in the WMT23 data task.

Nonetheless, the results remain meaningful, as we carefully controlled all experimental variables. 
BiMax slightly outperforms OT
while offering a noticeably faster processing speed~(as shown in Table~\ref{tab:info_benchmark}),
indicating that parallel sentence pair datasets generated using BiMax can match OT in quality 
while requiring fewer computational resources. 
Moreover, as described in Section~\ref{sec:dev_bench2},
document alignment holds strong potential for producing high-quality translation data. 
Simply appending a basic baseline dataset can enable performance 
that rivals—or even exceeds—that of more complex data construction pipelines designed by WMT23 participants.

As mentioned in Section~\ref{sec:intro}, 
the advancement of large language models (LLMs) has rendered document-level translation increasingly feasible. 
Therefore, rather than adhering to the conventional practice of evaluating alignment quality
using downstream sentence-level MT systems, 
it may be more effective to assess document alignment directly through document-level machine translation.
\end{document}